\documentclass[sigconf]{acmart}

\usepackage[utf8]{inputenc} 
\usepackage[T1]{fontenc}    
\usepackage{hyperref}       
\usepackage{url}            
\usepackage{booktabs}       
\usepackage{amsfonts}       
\usepackage{nicefrac}       
\usepackage{microtype}      
\usepackage{amsmath}
\usepackage{color}
\usepackage{bm}
\usepackage{subfigure}
\usepackage{graphicx}
\usepackage{algorithm}
\usepackage{algorithmic}
\usepackage{natbib}
\setcopyright{rightsretained}

\copyrightyear{2021}
\acmYear{2021}
\setcopyright{acmlicensed}\acmConference[GECCO '21]{2021 Genetic and Evolutionary Computation Conference}{July 10--14, 2021}{Lille, France}
\acmBooktitle{2021 Genetic and Evolutionary Computation Conference (GECCO '21), July 10--14, 2021, Lille, France}
\acmPrice{15.00}
\acmDOI{10.1145/3449639.3459292}
\acmISBN{978-1-4503-8350-9/21/07}

\begin{document}

\title{Regularized Evolutionary Population-Based Training}


\settopmatter{authorsperrow=4}
\author{Jason Liang}
\affiliation{Cognizant AI Labs\\San Francisco, CA, USA}
\author{Santiago Gonzalez}
\affiliation{Cognizant AI Labs\\\hspace*{-0.5ex}The Univ.\ of Texas at Austin}
\author{Hormoz Shahrzad}
\affiliation{Cognizant AI Labs\\\hspace*{-0.5ex}The Univ.\ of Texas at Austin}
\author{Risto Miikkulainen}
\affiliation{Cognizant AI Labs\\\hspace*{-0.5ex}The Univ.\ of Texas at Austin}

\begin{abstract}
Metalearning of deep neural network (DNN) architectures and hyperparameters has become an increasingly important area of research. At the same time, network regularization has been recognized as a crucial dimension to effective training of DNNs. However, the role of metalearning in establishing effective regularization has not yet been fully explored. There is recent evidence that loss-function optimization could play this role, however it is computationally impractical as an outer loop to full training. This paper presents an algorithm called Evolutionary Population-Based Training (EPBT) that interleaves the training of a DNN's weights with the metalearning of loss functions. They are parameterized using multivariate Taylor expansions that EPBT can directly optimize. Such simultaneous adaptation of weights and loss functions can be deceptive, and therefore EPBT uses a quality-diversity heuristic called Novelty Pulsation as well as knowledge distillation to prevent overfitting during training. On the CIFAR-10 and SVHN image classification benchmarks, EPBT results in faster, more accurate learning. The discovered hyperparameters adapt to the training process and serve to regularize the learning task by discouraging overfitting to the labels. EPBT thus demonstrates a practical instantiation of regularization metalearning based on simultaneous training. 
\end{abstract}

%
%

\ccsdesc[500]{Computing methodologies~Neural networks}
\ccsdesc[300]{Computing methodologies~Distributed artificial intelligence}
\ccsdesc[100]{Computing methodologies~Computer vision}
\keywords{Deep Learning; Evolutionary Algorithms; Metalearning; Neural Networks; Regularization}

\maketitle

\section{Introduction}

Training modern deep neural networks (DNNs) often requires extensive tuning, and many seminal architectures have been developed through a hand-design process that requires extensive expertise \cite{lecun2015deep,NIPS2012_4824, bahdanau2014neural, mnih2015human}. To make this process easier and more productive, automated methods for metalearning and optimization of DNN hyperparameters and architectures have recently been developed, using techniques such as Bayesian optimization, reinforcement learning, and evolutionary search \cite{snoek2015scalable, zoph2016neural, miikkulainen2019evolving}. At the same time, regularization during training has become an important area of research, as preventing overfitting has been identified as crucial to the generalization capabilities of DNNs \cite{srivastava2014dropout, kukavcka2017regularization}. An interesting opportunity is therefore emerging: Using metalearning to discover improved regularization mechanisms \cite{cubuk2018autoaugment, balaji2018metareg, real2019regularized}. 

One promising such approach is evolution of loss functions. Instead of optimizing network structure or weights, evolution is used to modify the gradients, making it possible to automatically regularize the learning process \cite{houthooft2018evolved, gonzalez2019glo}. However, as in most of the earlier metalearning methods, evolution serves as an outer loop to network training. Such an approach is computationally prohibitive since fitness evaluations in principle need full training of DNNs. Also, the approach cannot adapt loss functions to different stages of learning.

Population-Based Training (PBT) was recently proposed to overcome this limitation in metalearning \cite{jaderberg2017population}. PBT interleaves DNN weight training with the optimization of hyperparameters that are relevant to the training process but also have no particular fixed value (e.g., learning rate). Such online adaptation is crucial in domains where the learning dynamics are non-stationary. Therefore, PBT forms a promising starting point for making loss-function optimization practical as well.

Building on PBT, this paper develops \textit{Evolutionary Population-Based Training} (EPBT) as such an approach through four extensions. First, powerful heuristics from evolutionary black-box optimization are employed to discover promising combinations of hyperparameters for DNN training. In particular, EPBT uses selection, mutation, and crossover operators adapted from genetic algorithms \cite{whitley1994genetic} to find good solutions. 

Second, a recently developed loss function parameterization based on multivariate Taylor expansions called \textit{TaylorGLO} \cite{gonzalez2020taylorglo} is combined with EPBT to optimize loss functions. This parameterization makes it possible to encode many different loss functions compactly and works on a variety of DNN architectures. In prior work, it was found to discover loss functions that result in faster training and better convergence than the standard cross-entropy loss.

Third, an interesting challenge emerges when both loss functions and the weights are adapted at the same time: The problem becomes inherently deceptive. Configurations that allow for fast learning in the beginning are often bad for fine tuning at the end of training. To overcome this problem, EPBT makes use of Novelty Pulsation \cite{shahrzad2018noveltyselection, shahrzad2020noveltypulsation}, a powerful heuristic for maintaining population diversity and escaping from deceptive traps during optimization. 

Fourth, another challenge with the coadaptation is that training is noisy and can overfit to the validation dataset during evolution.  EPBT thus introduces a new variant of knowledge distillation \cite{hinton2015distilling} called Population-Based Distillation. This method stabilizes training, helps reduce evaluation noise, and thus allows evolution to make more reliable progress.

Experiments on the CIFAR-10 and SVHN image classification benchmarks with several network architectures show that EPBT results in faster, more accurate learning. An analysis of the shapes of the discovered loss functions suggests that they penalize overfitting, thus regularizing the learning process automatically. Different loss shapes are most effective at different stages of the training process, suggesting that an adaptive loss functions perform better than one that remains static throughout the training, thus taking advantage of the synergy between learning and evolution.

\section{Background and Related Work}

This section summarizes relevant work in metalearning and regularization of DNNs, especially through population-based methods. Previous work on loss function optimization and novelty pulsation is also reviewed.

\subsection{Metalearning}

Metalearning of good DNN hyperparameters and architectures is a highly active field of research. One popular approach is to use reinforcement learning to tune a controller that generates the designs \cite{zoph2016neural, zoph2018learning, pham2018efficient}. Another approach is to make the metalearning differentiable to the performance of the DNN, and then learn by gradient descent \cite{maclaurin2015gradient, liu2018darts}.

Recently, metalearning methods based on evolutionary algorithms (EA) have also gained popularity. These methods can optimize DNNs of arbitrary topology and structure \cite{miikkulainen2019evolving}, achieving state-of-the-art results e.g.\ on large-scale image classification benchmarks \cite{real2019regularized}, and demonstrating good trade-offs in multiple objectives such as performance and network complexity \cite{lu2018nsga}. Many of these EAs use proven and time-tested heuristics such as mutation, crossover, selection, and elitism \cite{goldberg1988genetic, whitley1994genetic, nature_neuroevolution} to perform black-box optimization on arbitrary complex objectives. Advanced EAs such as CMA-ES \cite{hansen1996cmaes} have also optimized DNN hyperparameters successfully in high-dimensional search spaces \cite{loshchilov2016cma} and are competitive with statistical hyperparameter tuning methods such as Bayesian optimization \cite{snoek2012practical, snoek2015scalable, klein2017fast}. 

\subsection{Population-based Training}

One challenge shared by every DNN metalearning algorithm is deciding the right amount of training required to evaluate a network architecture and hyperparameter configuration on a benchmark task. Many algorithms simply stop training prematurely, assuming that the partially trained performance is correlated with the true performance \cite{li2017hyperband, miikkulainen2019evolving}. Other methods rely on weight sharing, where many candidate architectures share model layers \cite{pham2018efficient}, thus ensuring that the training time is amortized among all solutions being evaluated. 

PBT \cite{jaderberg2017population} uses the weight sharing approach, which is more computationally efficient. PBT works by alternating between training models in parallel and tuning the model's hyperparameters through an exploit-and-explore strategy. During exploitation, the hyperparameters and weights of well-performing models are duplicated to replace the worst performing ones. During exploration, hyperparameters are randomly perturbed within a constrained search space. Because PBT never retrains models from scratch, its computational complexity scales only with the population size and not with the total number of hyperparameter configurations searched. Besides tuning training hyperparameters such as the learning rate, PBT has successfully discovered data augmentation schedules \cite{ho2019population}. Therefore, PBT serves as a promising basis for the design of EPBT.

\subsection{Regularization of DNNs}

While metalearning seeks to find good DNN architectures, regularization is concerned about preventing DNNs from overfitting during training or optimization. Besides classic penalty-based approaches such as weight decay \cite{moody1995simple}, there are methods that leverage the structure of DNN layers. One simple but popular approach is dropout \cite{srivastava2014dropout}, which randomly sets the outputs of a layer to zero. This approach helps prevent overfitting by forcing subsequent layers to adapt to the noise generated by the previous layers. A related technique is batch normalization \cite{ioffe2015batch}, which normalizes the outputs of layers and prevents exploding gradients. These approaches work universally on all most all DNN architectures and problem domains and can even be combined.

Recently, much attention has been focused on manipulating training data to help regularize network training. Advanced data augmentation techniques such as cutout \cite{devries2017improved}, and cutmix \cite{yun2019cutmix} purposely create more diverse distributions of the input data to improve generalization and avoid overfitting. Similarly, adversarial examples \cite{goodfellow2014explaining} are another way to regularize by training the network with inputs that are particularly difficult for it to get right. Techniques such as label smoothing \cite{muller2019does} and knowledge distillation/self-distillation \cite{hinton2015distilling, kim2020self} soften the training targets to ensure more properly behaved gradients, resulting in better generalization. Along the same lines, loss-function optimization is an alternative method to achieve better regularization through modification of gradients, as will be reviewed next.

\subsection{Loss Function Optimization}

DNNs are trained through the backpropagation of gradients that originate from a loss function \cite{lecun2015deep}. Loss functions represent the primary training objective for a neural network. The choice of the loss function can have a significant impact on a network's performance \cite{janocha2017loss, bosman2019visualising, gonzalez2019glo}. Recently, Genetic Loss Optimization (GLO) \cite{gonzalez2019glo} was proposed as a new kind of metalearning, making it possible to automatically discover novel loss functions that can be used to train higher-accuracy neural networks in less time.

In GLO, loss functions are represented as trees and optimized through genetic programming \cite{banzhaf1998genetic}. This approach has the advantage of allowing arbitrarily complex loss functions. However, there are pathological functions in this search space with undesirable behaviors, such as discontinuities. To avoid these issues, this paper uses a loss-function representation based on multivariate Taylor expansions \cite{gonzalez2020taylorglo}. TaylorGLO parameterization is smoother, has guaranteed continuity and adjustable complexity, and is easier to implement. How these loss functions are optimized with EPBT is described in the next section.

\subsection{Novelty Selection and Pulsation}
\label{sc:noveltyselection}

Novelty Selection is a form of novelty search that augments the original fitness-based selection with novelty and thus improves the quality-diversity of the population \cite{lehman2008exploiting, shahrzad2018noveltyselection}. Novelty is measured through a behavioral description of the individuals, i.e.\ a phenotypical feature vector that is not related to fitness. An initial set of $m$ elite candidates is first selected based on fitness and sorted according to their novelty score $S_i$, measured as the sum of pairwise distances $d$ of the individual's behavior vector $b_i$ to those of all other individuals $j$ in the set:
\begin{equation}
\label{eq-NoveltyScore}
S_{i} = \sum_{j = 1}^{m}d(b_{i},\ b_{j})\; .
\end{equation}
The top $k$ candidates from this set are selected as elites, skipping candidates that represent the same cluster.

In principle, fitness-based selection is more greedy than novelty selection, and could result in faster convergence. On the other hand, novelty selection explores more diverse candidates, which could help discover better regularization. In Novelty Pulsation \cite{shahrzad2020noveltypulsation}, such exploitation and exploration are both leveraged by switching Novelty Selection on and off for every $p$ generations, resulting in faster convergence and more reliable solutions. Novelty Pulsation plays an important role in keeping the population diverse enough to avoid deceptive interactions between weight and loss-function adaptation in EPBT. Furthermore, unlike other novelty-based methods such as Map Elites \cite{mouret2015illuminating}, Novelty Pulsation can be easily integrated with EPBT due to the simplicity of the algorithm and lack of computational overhead from maintaining an archive of individuals.

\section{The EBPT Method}\label{sec:epbt}

This section describes in detail how EPBT implements regularization metalearning. EPBT utilizes genetic operators from black-box optimization to enhance hyperparameter metalearning in PBT, and combines it with loss-function metalearning. Their deceptive and overfitting interactions are mitigated through a selection heuristic based on quality-diversity, and through knowledge distillation.

\subsection{Overview}

\begin{figure}[t]
 \begin{center}
   \includegraphics[width=\linewidth]{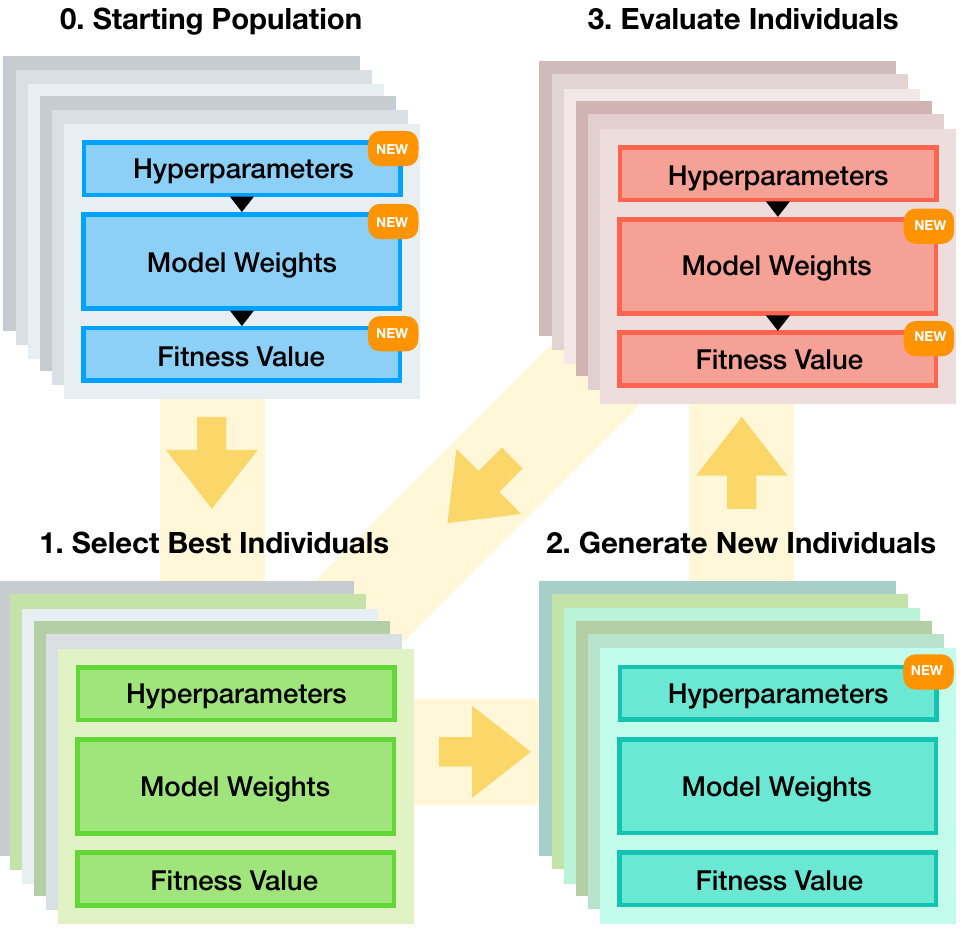}
   \caption{An overview of the EPBT evolutionary loop. EPBT begins by randomly initializing individuals, which are composed of hyperparameters, model weights, and fitness values. Next, EPBT runs for multiple generations in a three-step loop: (1) selection of the best individuals, (2) generation of new individuals, and (3) evaluation of these individuals. In Step 1, promising individuals are selected using a heuristic. In Step 2, new individuals with updated hyperparameters are created, but the weights and fitness are inherited. In Step 3, these individuals are evaluated on a task and have their model weights and fitness (i.e., performance in the task) updated. Thus, EPBT makes it possible to simultaneously train the network and evolve loss function parameterizations.}
   \label{fig:epbt}
 \end{center}
\vspace*{-2ex}
\end{figure}

The core concept of EPBT is intuitive and builds on extensive work already done with evolutionary optimization of DNNs \cite{nature_neuroevolution}. Figure~\ref{fig:epbt} shows how EPBT maximizes the fitness of a population of candidate solutions (individuals) over multiple iterations (generations). As a black-box method, EPBT requires no gradient information but only the fitness value of each individual. With EPBT, it is thus possible to apply metalearning to tasks where meta-gradients are not available.

At the beginning of generation $g$, the population $\mathbb{M}_g$ consists of individuals $M_{gi}$. Each $M_{gi} = \{\mathbf{D}_{gi}, \mathbf{h}_{gi}, f_{gi}\}$ where $\mathbf{D}_{gi}$ is a DNN model (defined as the weights of an user-specified architecture), $\mathbf{h}_{gi}$ is a set of hyperparameters, and $f_{gi}$ is a real-valued scalar fitness. In the Step 1 of the generation, $f_{gi}$ is used to select promising individuals $\hat{M}_{gi}$ to form a parent set $\mathbb{\hat{M}}_g$, where $\mathbb{\hat{M}}_g \subset \mathbb{M}_g$. In the Step 2, $\mathbb{\hat{M}}_g$ is used to create a set $\mathbb{N}_g$, which contains new individuals $N_{gi}$. Each of these new individuals inherits $\mathbf{D}_{gi}$ unchanged from its parent $\hat{M}_{gi}$, but with modified hyperparameters $\mathbf{\hat{h}}_{gi}$. The genetic operators used for generating $\mathbb{N}_g$ will be described in more detail in the next subsection. Finally, in Step 3, each $N_{gi}$ is evaluated by training $\mathbf{D}_{gi}$ on a task or dataset, thereby creating a model with updated weights $\mathbf{\hat{D}}_{gi}$. The validation performance of $\mathbf{\hat{D}}_{gi}$ is used to determine a new fitness value $\hat{f}_{gi}$. Thus, by the end of generation $g$, the population pool contains the evaluated individuals $\hat{N}_{gi} \in \mathbb{\hat{N}}_g$, where $\hat{N}_{gi} = \{\mathbf{\hat{D}}_{gi}, \mathbf{\hat{h}}_{gi}, \hat{f}_{gi}\}$. This process is repeated for multiple generations until the fitness of the best individual in the population converges.

Within the core metalearning evolutionary loop, EPBT contains several other components. They include: (1) A collection of genetic operators specifically chosen for the task of hyperparameter optimization. (2) The Novelty Selection and Pulsation search heuristic that improves population diversity by preserving the most novel elites. (3) A TaylorGLO representation of loss functions with parameters that EPBT can optimize. (4) Population-Based Distillation method (PBD) that uses the best model in the population to help train other networks. Each of these components will be described in more detail below. 

Note that since the evaluation of an individual does not depend on other individuals, the entire EPBT process can be parallelized. In the current implementation of EPBT, fitness evaluations are mapped onto a multi-process pool of workers on a single machine. Each worker has access to a particular GPU of the machine, and if there are multiple GPUs available, every GPU will be assigned to at least one worker. For the experiments in this paper, a single worker does not fully utilize the GPU and multiple workers can be trained in parallel without any slowdown.  

\subsection{Genetic Operators}

EPBT uses standard evolutionary black-box optimization operators \cite{whitley1994genetic} to tune individuals. This subsection details how EPBT is initialized and how these operators are utilized through the three septs of each generation, with a summary in Algorithm~\ref{alg:epbt}.

\begin{algorithm}[t]
   \caption{EPBT}
   \label{alg:epbt}
   \begin{algorithmic}
   \STATE {\bfseries Input:} max generations $n$, initial population $\mathbb{M}_0$, genetic operators $\tau, \gamma, \xi$
   \FOR{$g=0$ {\bfseries to} $n-1$}
       \STATE 1. Select $\hat{M}_{gi} = \{\mathbf{D}_{gi}, \mathbf{h}_{gi}, f_{gi}\}$ using $\tau$
       \STATE 2a. Set $\mathbf{\hat{h}}_{gi} = \xi(\gamma(\mathbf{h}_{gi})))$
       \STATE 2b. Set $N_{gi} = \{\mathbf{D}_{gi}, \mathbf{\hat{h}}_{gi}\}$
       \STATE 3a. Evaluate $\mathbb{N}_g$, set $\hat{N}_{gi} = \{\mathbf{\hat{D}}_{gi}, \mathbf{\hat{h}}_{gi}, \hat{f}_{gi}\}$
       \STATE 3b. Set $\mathbb{\dot{M}}_g$ to top $k$ $M_{gi}$ from $\mathbb{M}_g$
       \STATE 3c. Set $\mathbb{M}_{g+1} = \mathbb{\hat{N}}_g \cup  \mathbb{\dot{M}}_g$
   \ENDFOR
   \end{algorithmic}
\end{algorithm}

\textbf{Initialization:} A population with $P$ individuals is created as $\mathbb{M}_0$. For each $M_{0i} \in \mathbb{M}_0$, $\mathbf{D}_{0i}$ is set to a fixed DNN architecture and its weights are randomly initialized. Also, each variable in $\mathbf{h}_{0i}$ is uniformly sampled from within a fixed range and $f_{0i}$ is set to zero. 

\textbf{Step 1 -- Tournament Selection:} Using the tournament selection operator $\tau$, $t$ individuals are repeatedly chosen at random from $\mathbb{M}_g$. Each time, the individuals are compared and the one with the highest fitness is added to $\mathbb{\hat{M}}_g$. This process is repeated until $|\mathbb{\hat{M}}_g| = |\mathbb{M}_g| - k$, where $k$ is the number of elites. The value $t=2$ is commonly used in EA literature and in the experiments in this paper also.

\textbf{Step 2 -- Mutation and Crossover:} For each $\hat{M}_{gi}$, a uniform mutation operator $\gamma$ is applied by introducing multiplicative Gaussian noise independently to each variable in $\mathbf{h}_{gi}$. The mutation operator can randomly and independently reinitialize every variable as well. This approach allows for the exploration of novel combinations of hyperparameters. After mutation, a uniform crossover operator $\xi$ is applied, where each variable in $\mathbf{h}_{gi}$ is randomly swapped (50\% probability) with the same variable from another individual in $\mathbb{\hat{M}}_g$, resulting in the creation of $\mathbf{\hat{h}}_{gi}$. $\mathbf{D}_{gi}$ is copied from $\hat{M}_{gi}$ and combined with $\mathbf{\hat{h}}_{gi}$ to form the unevaluated individual $N_{gi}$.

\textbf{Step 3 -- Fitness Evaluation with Elitism:} The evaluation process proceeds as described above and results in evaluated individuals $\mathbb{\hat{N}}_g$. After evaluation, EPBT uses an elitism heuristic to preserve progress. In elitism without any Novelty Selection, $\mathbb{M}_g$ is sorted by $f_{gi}$ and the best $k$ performing individuals $\mathbb{\dot{M}}_g \subset \mathbb{M}_g$ are preserved and combined with $\mathbb{\hat{N}}_g$ to form $\mathbb{M}_{g+1}$, the population for the next generation. With Novelty Selection, the $k$ individuals are selected based on a combination of fitness and novelty, as was described in Section~\ref{sc:noveltyselection}. By default, $k$ is set to half of the population size, which is a popular value in literature. In the same way that mutation and crossover encourage exploration of a search space, elitism allows for the exploitation of promising regions in the search space.

\subsection{Loss Function Parameterization}

Loss functions are represented by leveraging the TaylorGLO parameterization. This parameterization is defined as a fixed set of continuous values, in contrast to the original GLO parameterization based on trees \cite{gonzalez2019glo}. TaylorGLO loss functions have several functional advantages over GLO: they are inherently more stable, smooth, and lack discontinuities \cite{gonzalez2020taylorglo}. Furthermore, because of their simple and compact representation as a continuous vector, TaylorGLO functions can be easily tuned using black-box methods. Specifically in this paper, a third-order TaylorGLO loss function with parameters $\theta_0 \ldots \theta_7$, is used:
\newcommand{\Tx}[1][-\theta_0]{(y_i #1)}
\newcommand{\Ty}[1][-\theta_1]{(\hat{y}_i #1)}
\begin{equation}
\scriptsize
\label{eq:k3taylorglo}
\begin{aligned}
\mathcal{L}(y,\hat{y}) = -\frac{1}{n}\sum^n_{i=1} \Big[    \theta_2\Ty + \frac{1}{2}\theta_3\Ty^2 + \frac{1}{6}\theta_4\Ty^3 + \\
+ \theta_5\Tx\Ty + \frac{1}{2}\theta_6\Tx\Ty^2 + \frac{1}{2}\theta_7\Tx^2\Ty  \Big] \;,
\end{aligned}
\end{equation}
where $y$ is the sample's true label in one-hot form, and $\hat{y}$ is the network's prediction (i.e., scaled logits). The eight parameters ($\theta_0 \ldots \theta_7$) are stored in $\mathbf{h}_{gi}$ and optimized using EPBT.

\subsection{Novelty Selection and Pulsation}

Novelty Pulsation works by turning Novelty Selection on and off at each pulsation cycle interval $p$, changing how elite individuals are chosen during Step 3. When Novelty Selection is on, $\mathbb{M}_g$ is first increased to include the most fit $m>k$ candidates, and then filtered down to $k$ most novel elites as described in Section~\ref{sc:noveltyselection}. For the experiments in this paper, the behavior metric used to compute novelty is a binary vector indicating whether the candidate correctly predicts the classes of a randomly chosen $N$-sized subset of the validation data. This behavior metric encourages evolution to discover models that can perform well generally and not just overfit to a few classes in particular. Preliminary experiments showed that setting $m = 3/2 k$, $N=400$, and $p=5$ works well and helps protect against premature convergence.

\subsection{Population-Based Distillation}

To avoid overfitting during evolution, model evaluation includes a variant of knowledge distillation \cite{hinton2015distilling}. The main idea in knowledge distillation is to construct training targets as a linear combination of sample labels and the predictions of a better performing teacher model. A model from the previous epoch can be used as a teacher, resulting in a strong regularizing effect \cite{kim2020self}. Thus, in PBD, the loss for an individual's model $\mathbf{D}_{gi}$ is computed as
\begin{equation}
\label{eq:pbd}
\mathcal{L}(y,\hat{y}) = \mathcal{L}(\hat{\alpha} * \hat{y}_p + (1 - \hat{\alpha}) * y, \hat{y}),
\end{equation}
where $\hat{y}_p$ are the predictions of the best individual in the previous generation. $\hat{\alpha} = \alpha * (t/T)$ where $t$ is the number of training epochs elapsed, $T$ the total number of training epochs, and $\alpha$ is a scalar between 0 and 1. As in prior work \cite{kim2020self}, to minimize the effect of inaccurate teacher models in the beginning, $\hat{\alpha}$ is initialized at zero and then increased linearly with the number of training epochs. Thus, PBD regularizes against overfitting to the validation dataset by smoothing the target labels and making for better-behaved gradients, allowing evolution to proceed more reliably.

\begin{figure}[t]
  \centering
  \includegraphics[width=\linewidth]{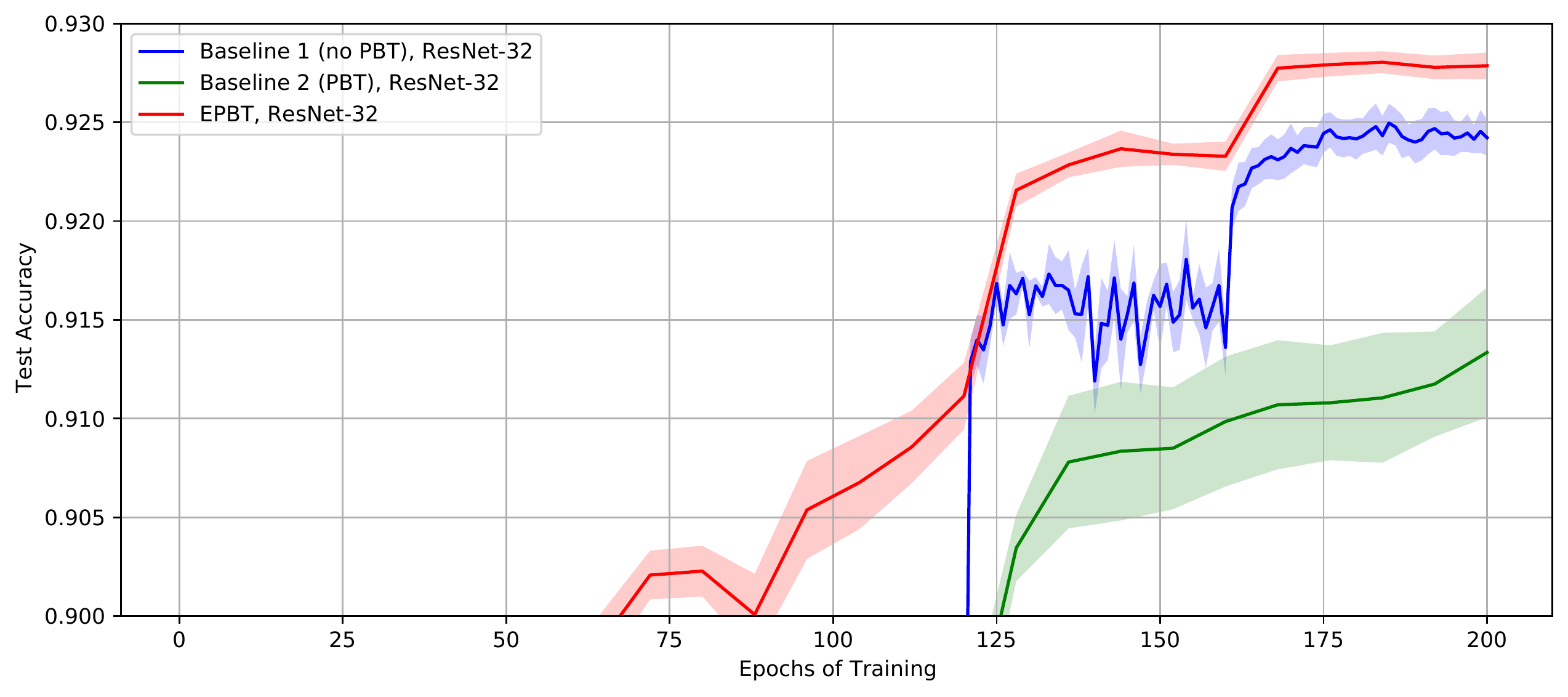}\\
  \includegraphics[width=\linewidth]{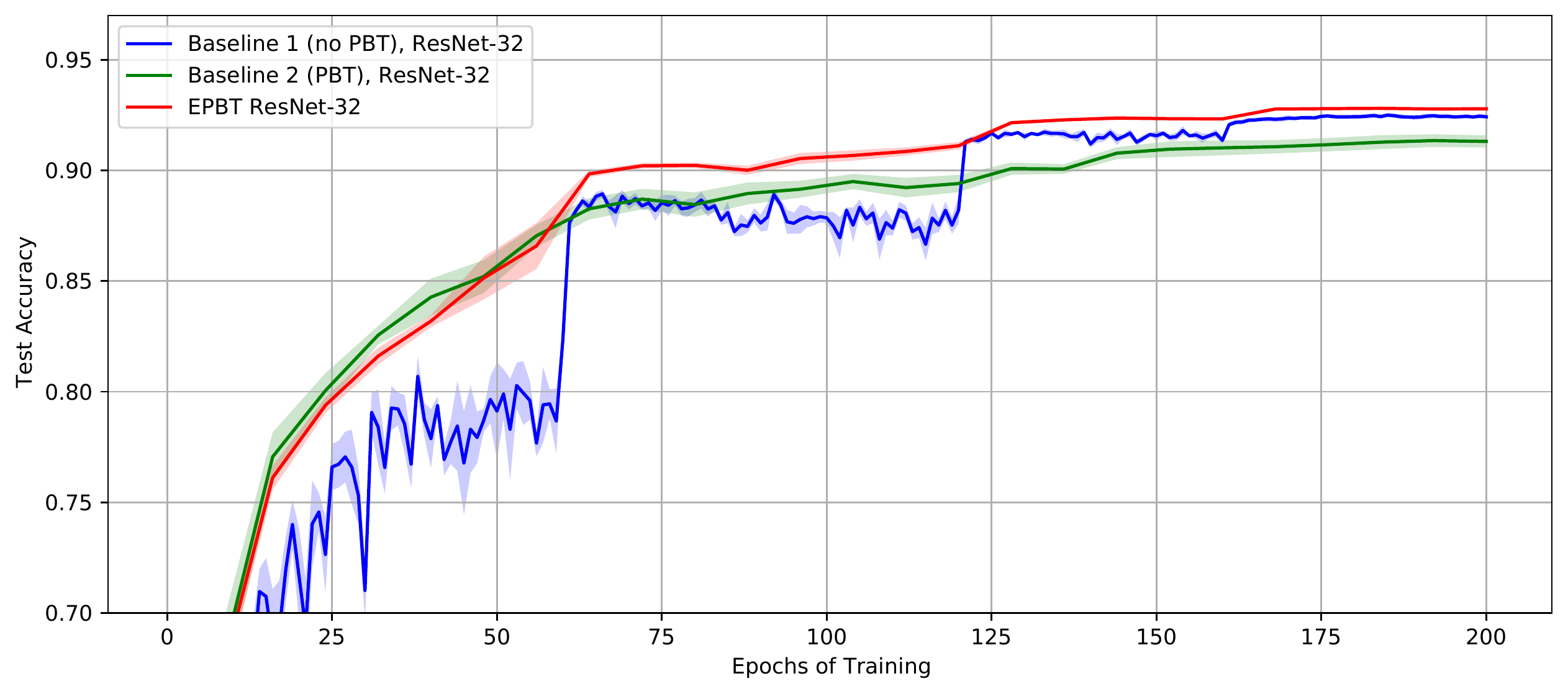}
  \caption{Experiments on CIFAR-10 with ResNet-32. Each line represents the test classification accuracy ($y$-axis) of the method over the number of epochs of training ($x$-axis). All results are averaged over five runs with error bars shown. The top plot is a zoomed-in version of the bottom plot. EPBT outperforms all baselines by a significant margin.}
  \label{fig:cifar10_res}
\end{figure}

\begin{figure*}[t]
  \centering
  \includegraphics[width=0.24\linewidth]{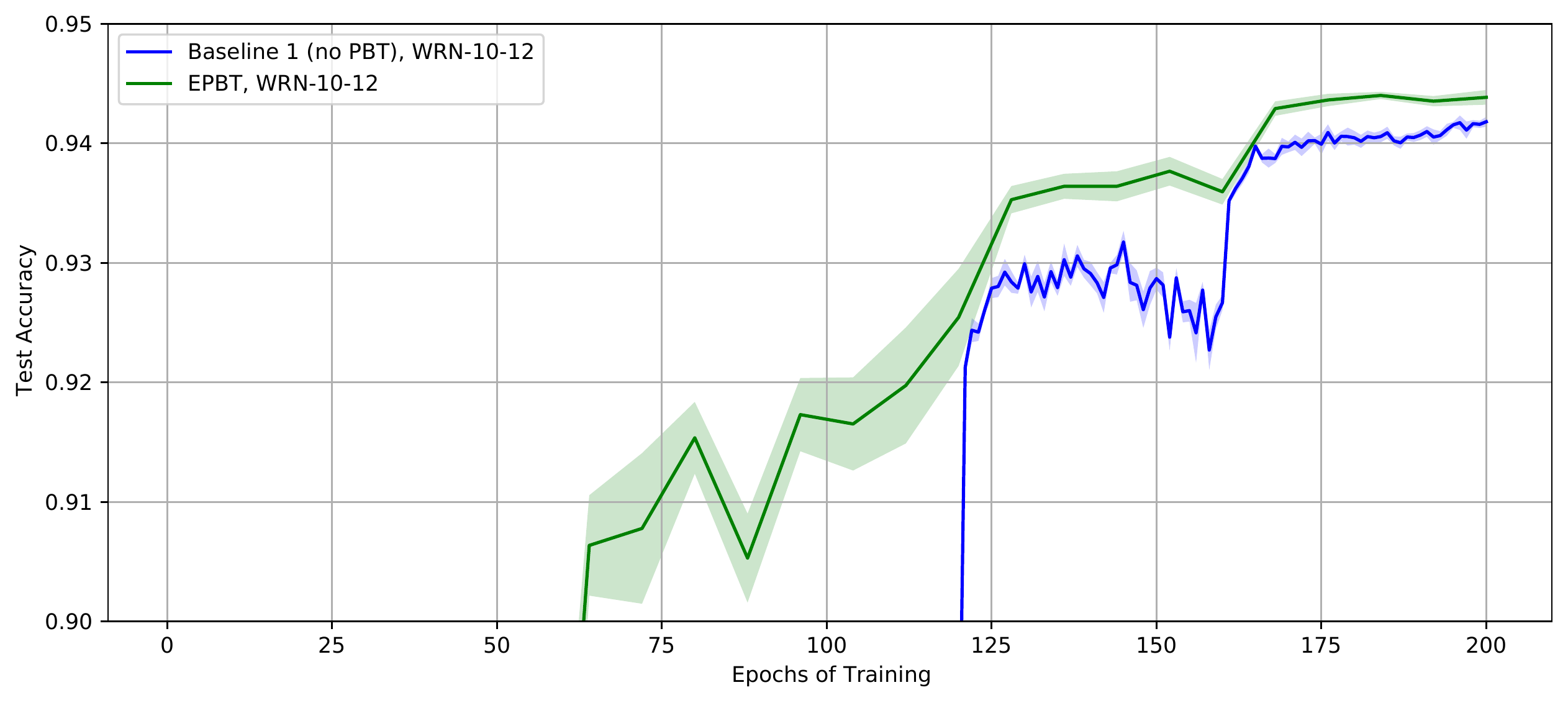}
  \includegraphics[width=0.24\linewidth]{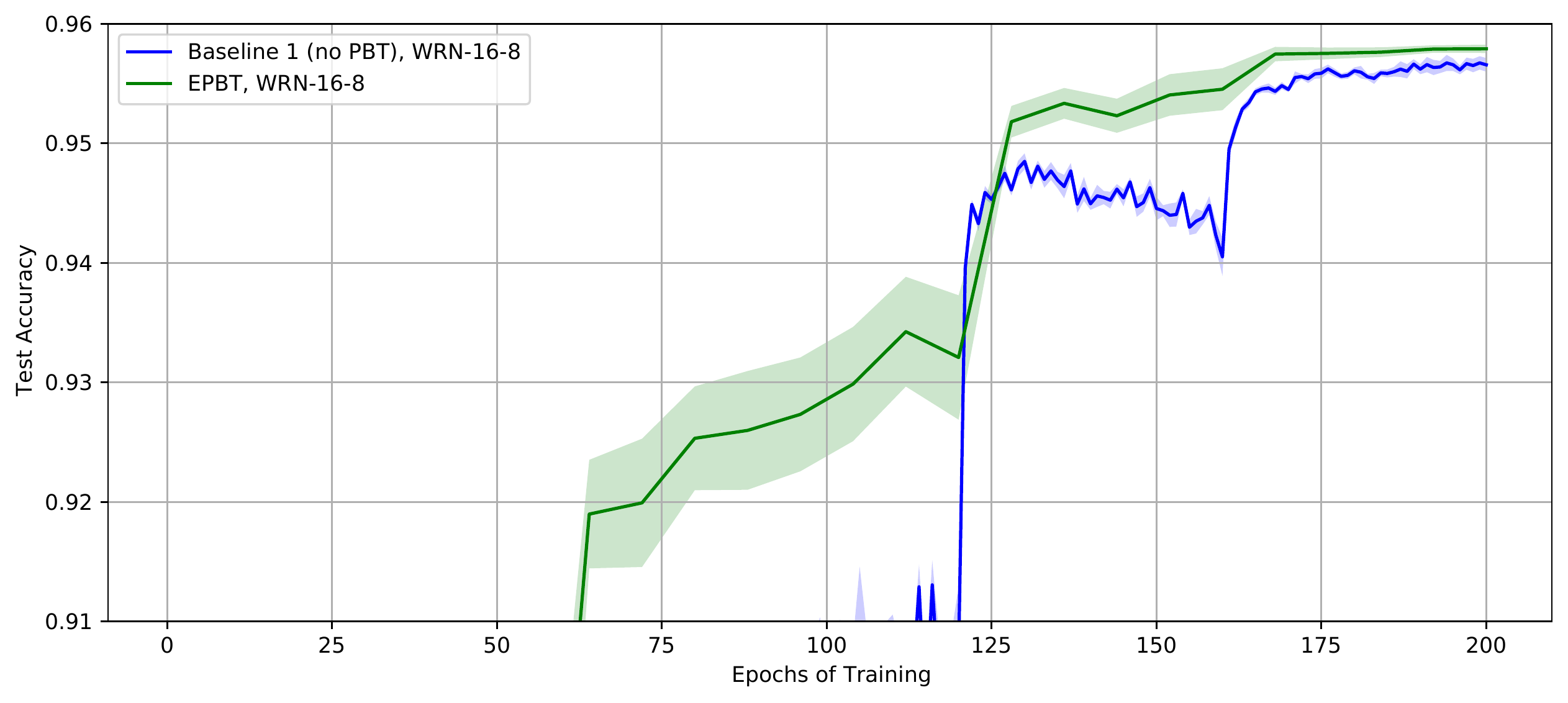}
  \includegraphics[width=0.24\linewidth]{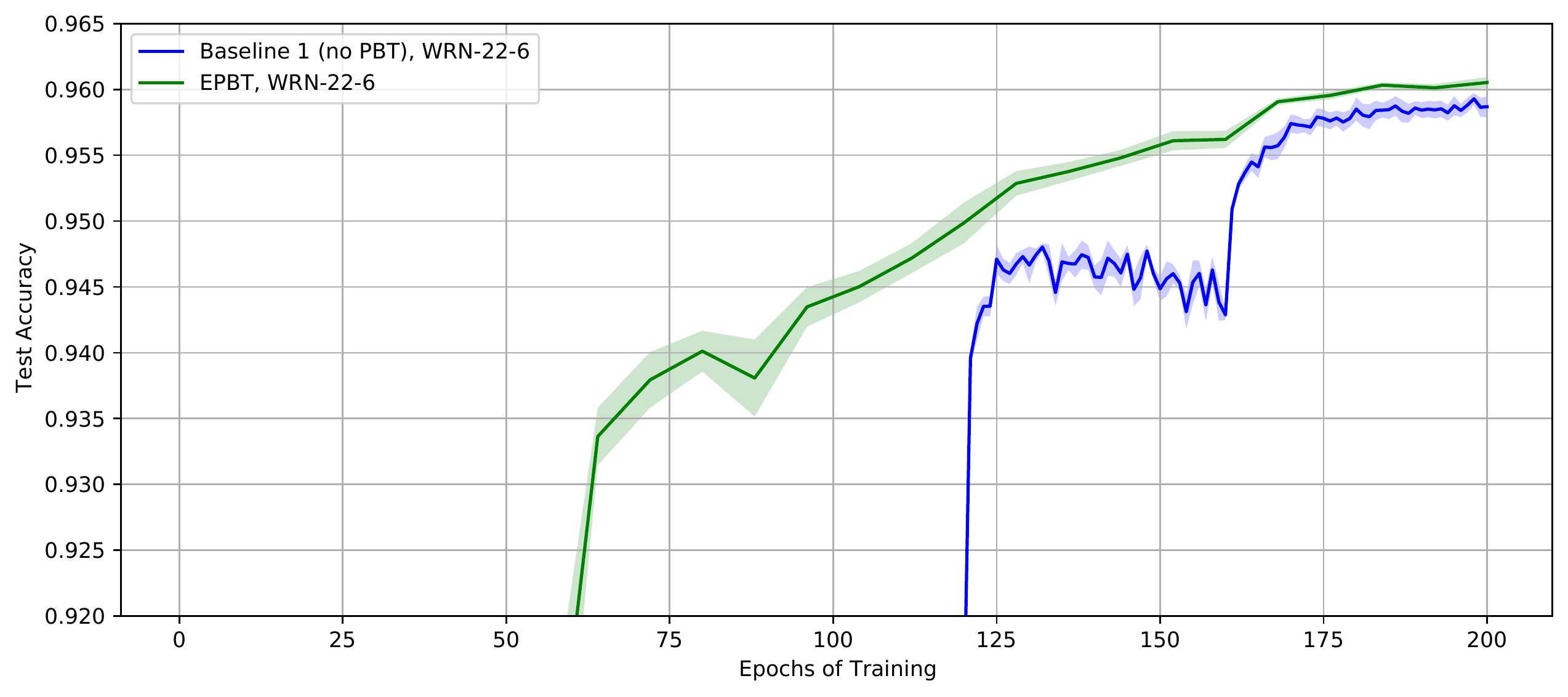}
  \includegraphics[width=0.24\linewidth]{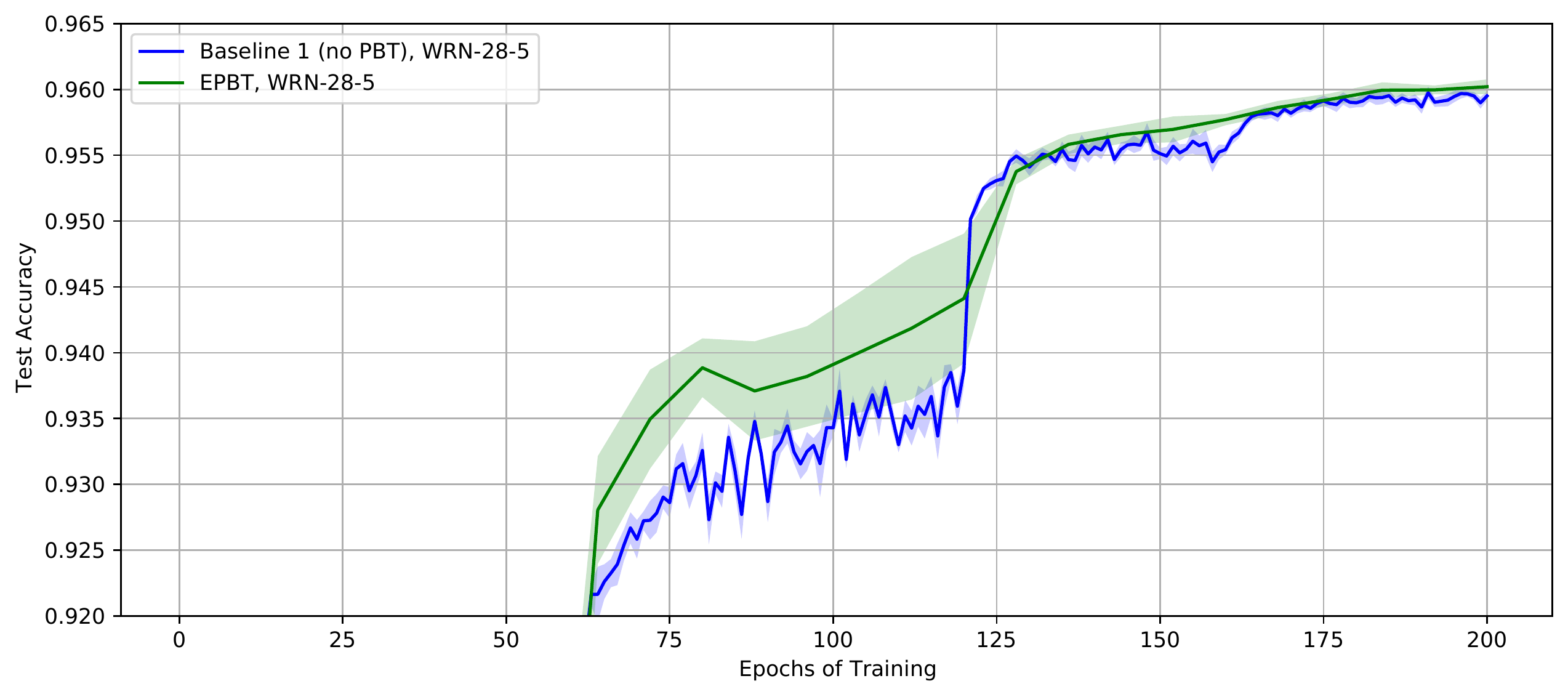}
  \includegraphics[width=0.24\linewidth]{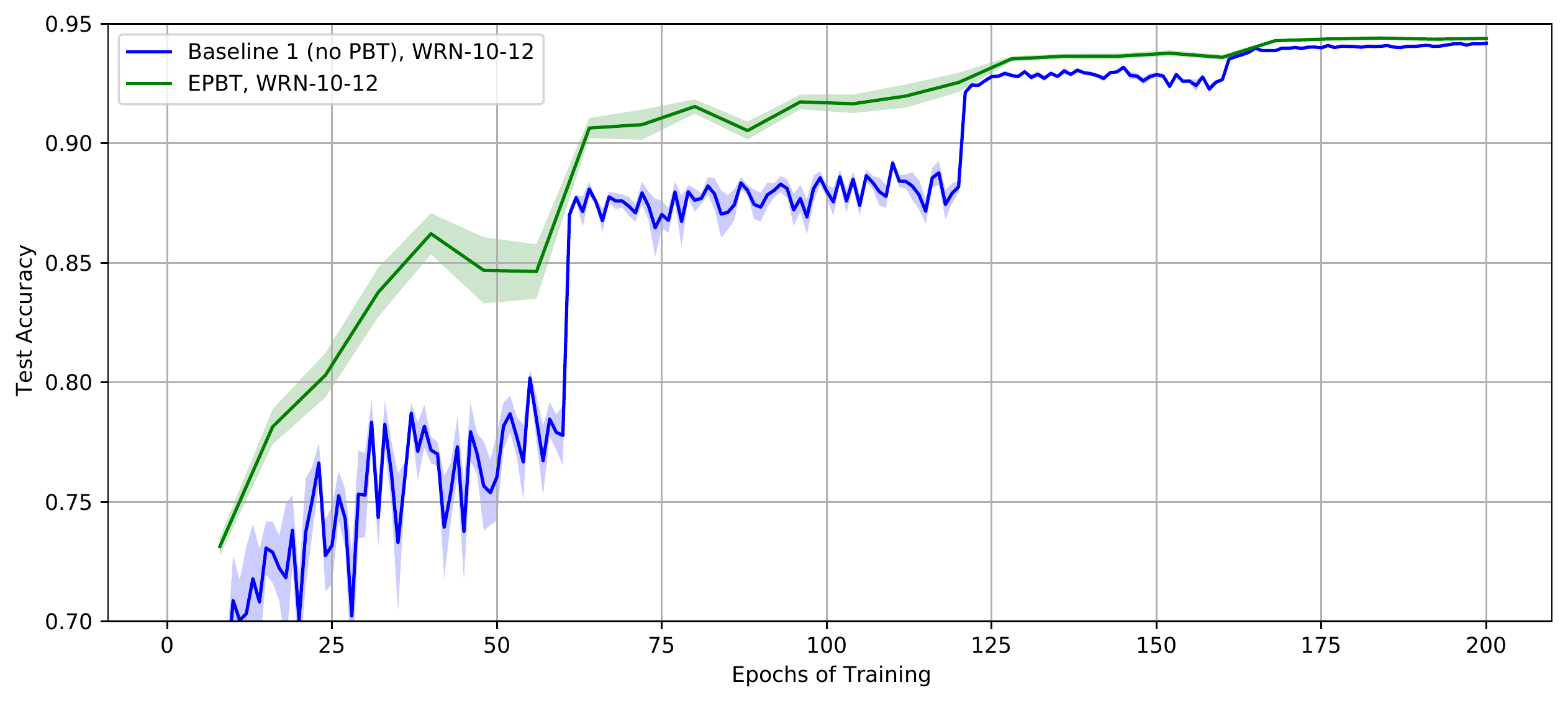}
  \includegraphics[width=0.24\linewidth]{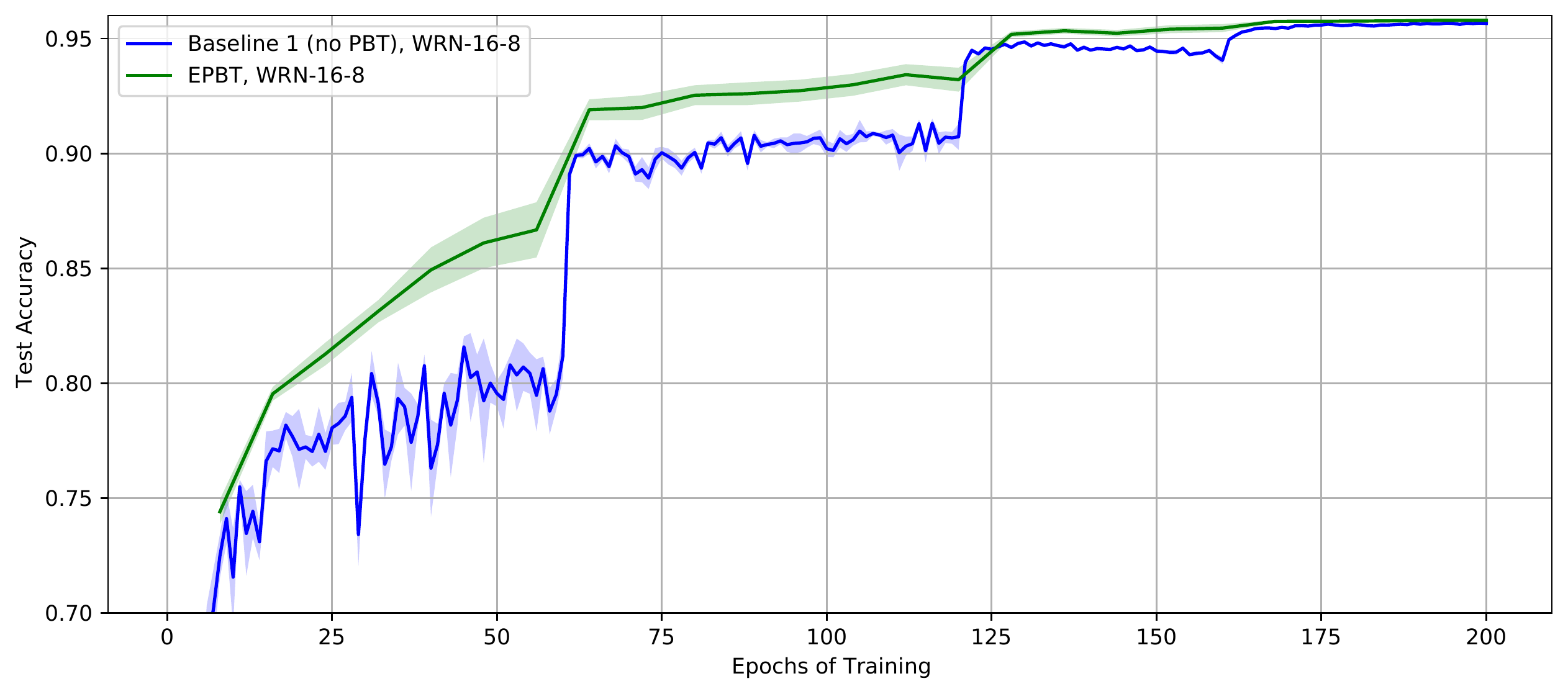}
  \includegraphics[width=0.24\linewidth]{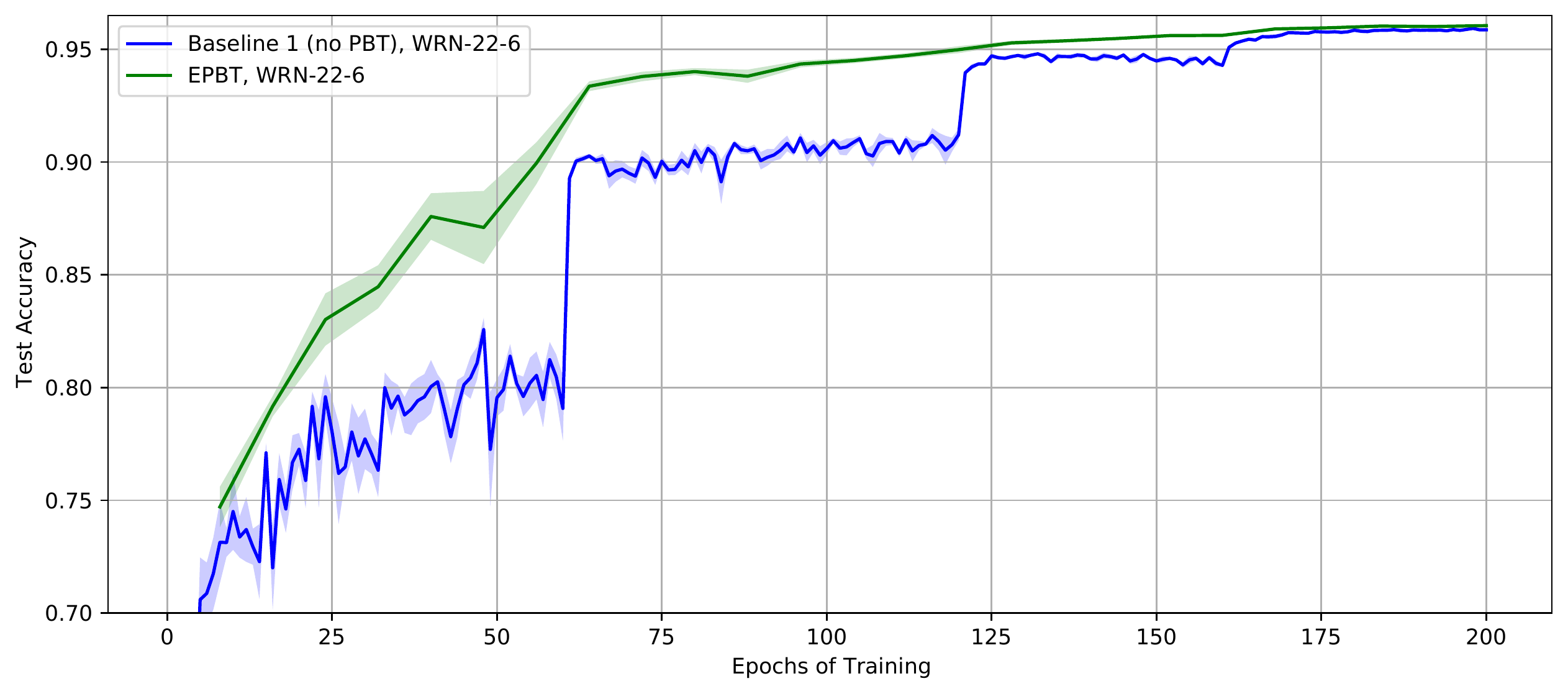}
  \includegraphics[width=0.24\linewidth]{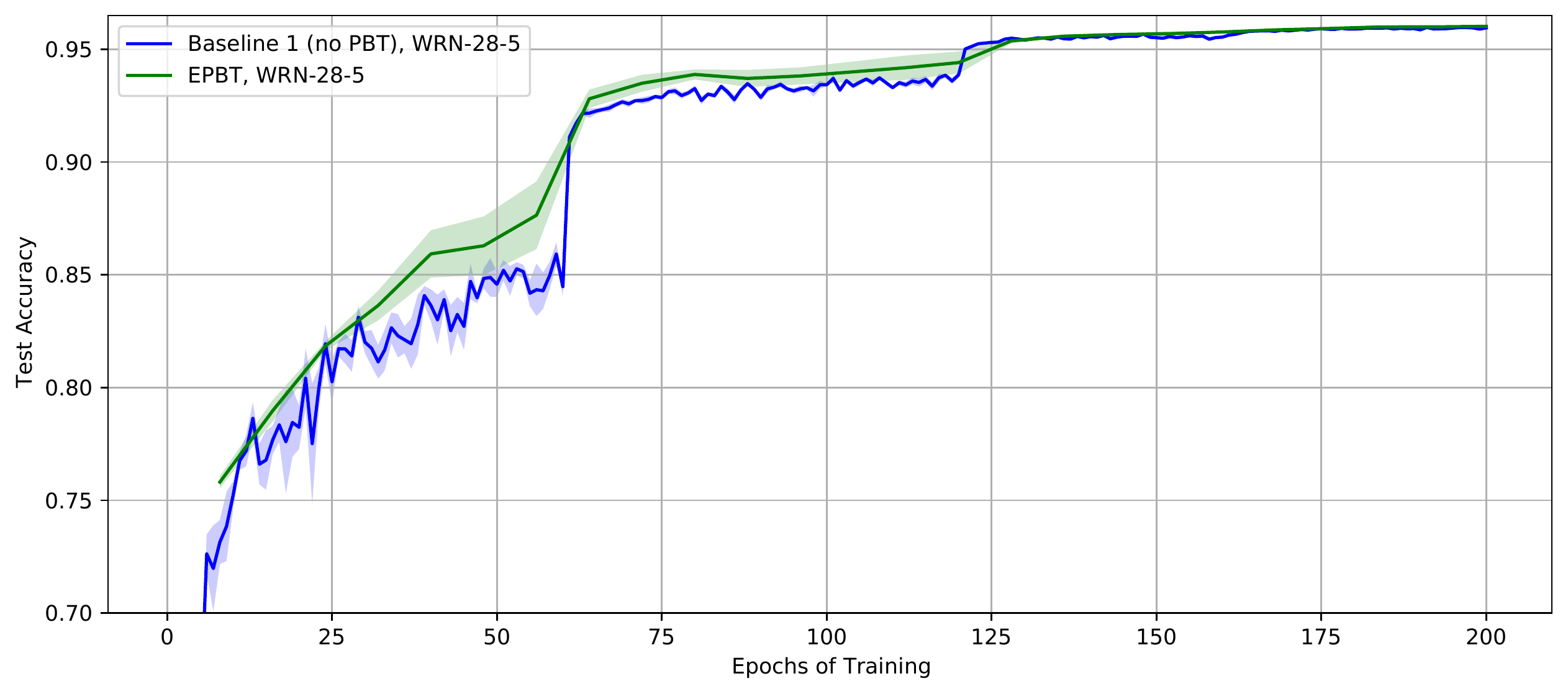}
  \caption{Experiments on CIFAR-10 with WRN-10-12, WRN-16-8, WRN-22-6, and WRN-28-5 (left to right). All results are averaged over five runs with error bars shown. The top plot is a zoomed-in version of the bottom plot. EPBT outperforms the baselines for all each of the architectures.}
  \label{fig:cifar10_wide}
\end{figure*}

\begin{figure}[t]
  \centering
   \includegraphics[width=\linewidth]{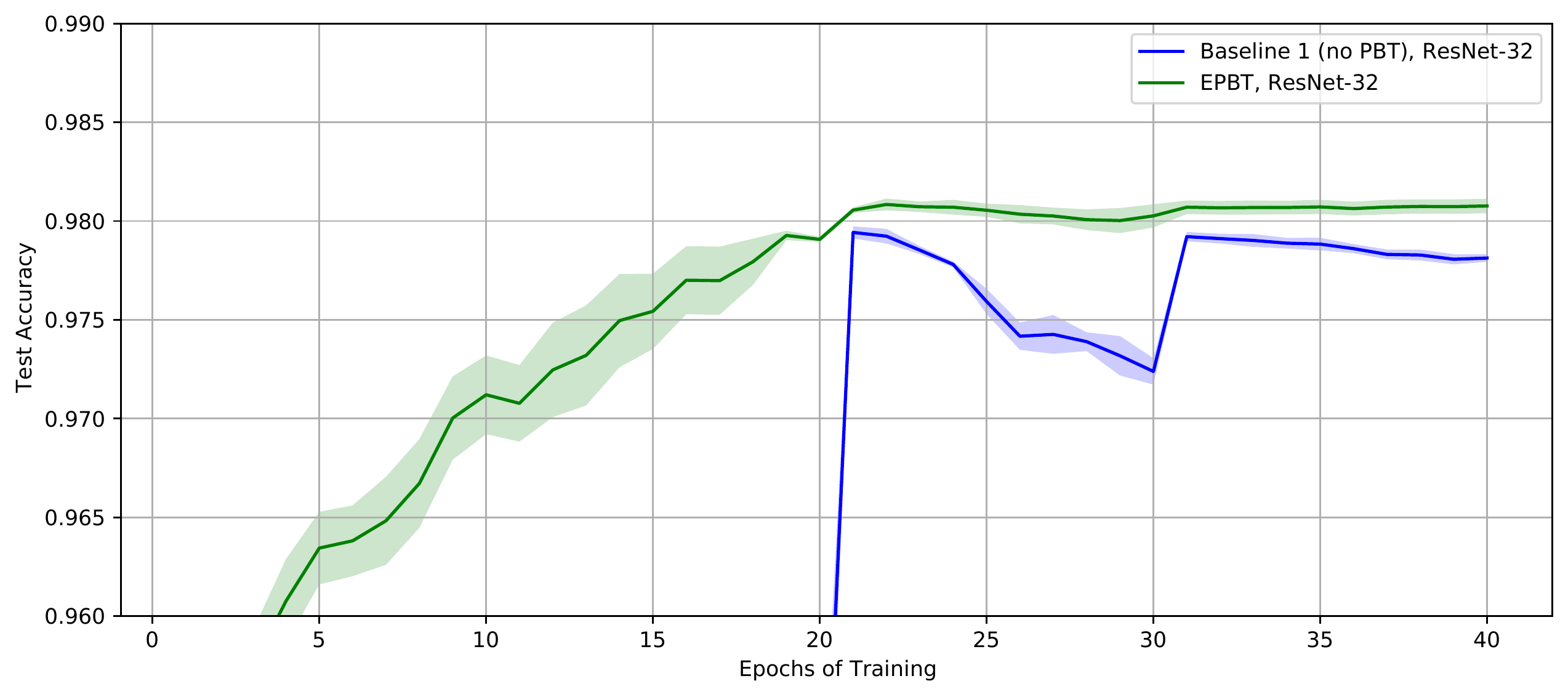}
   \includegraphics[width=\linewidth]{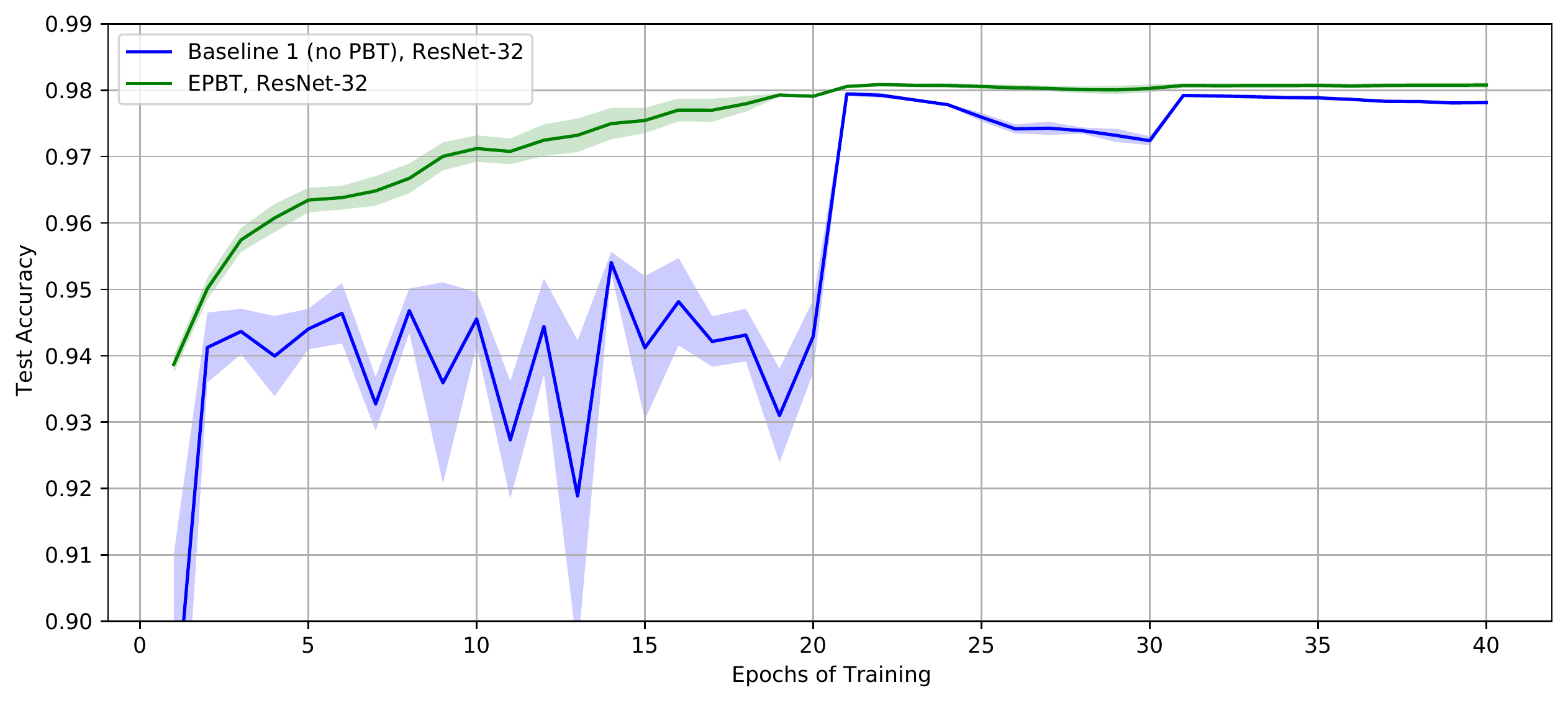}
  \caption{Experiments on SVHN with ResNet-32. All results are averaged over five runs with error bars shown. The top plot is a zoomed-in version of the bottom plot. EPBT outperforms the baseline, which uses cross-entropy loss to train.}
  \label{fig:svhn_res}
\end{figure}

\section{Experimental Results}

To show the effectiveness of EPBT, the algorithm was applied to optimize loss functions for two popular image classification datasets: CIFAR-10 and SVHN. Experimental results and comparisons to multiple baselines are presented below.

\subsection{CIFAR-10}

CIFAR-10 \cite{krizhevsky2009learning} is a widely used image classification dataset consisting of 60,000 natural images in ten classes. The dataset is composed of a training set of 50,000 images and a test set of 10,000 images. For evaluating individuals in EPBT, the training set was split into a separate validation set of 1,250 images and smaller training set with 48,750 images. To control noise during evaluations, the validation dataset was artificially enlarged to 25,000 images through data augmentation. The fitness was calculated by finding the classification accuracy of the trained model on this set. The test accuracies of each individual's model after evaluation were also recorded for comparison purposes only. 

To understand the improvements brought about by EPBT, two baselines were created. The first baseline is a model trained without EPBT: a 32-layer residual network (ResNet-32) with 0.47 million weights that was initialized with the He method \cite{he2015delving, resnet}. The model was trained using stochastic gradient descent (SGD) for 200 epochs on all 50,000 training images with a batch size of 128, momentum of 0.9, and cross-entropy loss. A fixed learning rate schedule that starts at 0.1 and decays by a factor of 5 at 60, 120, and 160 epochs was used. Input images were normalized to have unit pixel variance and a mean pixel value of zero before training while data augmentation techniques such as random flips, translations, and cutout \cite{devries2017improved} were applied during training.

The second baseline is a reimplementation of the original PBT algorithm that was used to optimize DNNs on the CIFAR-10 dataset \cite{jaderberg2017population}. The training setup was similar to the first baseline but with learning rate as an evolvable hyperparameter. Unlike EPBT, PBT only makes uses of truncation selection, where the weights and loss parameters from the top 25\% of the population were copied over to the bottom 25\% every generation, and simple mutation, where the hyperparameters were tuned using a mixture of both random resets and multiplicative perturbations of magnitude 1.2. PBT was run for 25 generations, each with eight epochs of training (for a total of 200 epochs), and with a population size of 40. The learning rate was randomly initialized between 0.1 and 0.0001. 

The experiments with EPBT were run using a similar training setup as described above. Like the PBT baselines, EPBT was run for 25 generations of eight epochs each and with a population size of 40. Besides the TaylorGLO loss function parameterization, EPBT optimized the hyperparameters for SGD learning rate schedule and momentum as well. The learning rate schedule was based on the one used by the first baseline but with a tunable scaling and decay factor. This search space allows for the exploration of novel schedules but can rediscover the original schedule if necessary. EPBT was configured similarly as the PBT baseline, but with an elitism size of $k=20$ and with the initial TaylorGLO parameters sampled uniformly between $-10$ and $10$. 

\begin{table*}[t]
  \begin{center}
  \begin{footnotesize}
  \begin{sc}
  \begin{tabular}{lcccccc}
  \toprule
  Algorithm & CIFAR-10, ResNet-32 & CIFAR-10, WRN-10-12 & CIFAR-10, WRN-16-8 &  CIFAR-10, WRN-22-6 & CIFAR-10, WRN-28-5 & SVHN, ResNet-32 \\
  \midrule
  Baseline 1 (no PBT) & 92.42 (0.21) & 94.18 (0.09) & 95.66 (0.13) & 95.87 (0.18) & 95.95 (0.07) & 97.81 (0.04) \\ 
  Baseline 2 (PBT) & 91.53 (0.56) & -- & -- & -- & -- & -- \\
  \midrule
  EPBT & \textbf{92.79 (0.15)} & \textbf{94.38 (0.14)} & \textbf{95.79 (0.08)} & \textbf{96.05 (0.09)} & \textbf{96.02 (0.12)} & \textbf{98.08 (0.08)} \\ 
  \bottomrule
  \end{tabular}
  \end{sc}
  \end{footnotesize}
  \end{center}
  \vskip 0.15in
  \caption{Mean and standard deviation (over five runs) of final test accuracies on the CIFAR-10 and SVHN datasets. EPBT achieves better results (bold) compared to the baselines. Results are reported in percentage.}
  \label{table:epbt_results}
  \vspace*{-3ex}
\end{table*}

The test accuracies of each baseline and best model in EPBT's population, averaged over five independent runs with standard error, are shown in Figure~\ref{fig:cifar10_res}. EPBT converges rapidly to the highest test accuracy and outperforms all other baselines. Baseline 2 (PBT) results in the worst performance, followed by the Baseline 1, which does not use a population at all. The relatively poor performance of PBT might be because it does not explicitly optimize for learning rate schedules. Previous work has shown that decaying the learning rate at crucial moments during training is important \cite{you2019does}. On the other hand, EPBT's more advanced parameterization and Novelty Pulsation heuristic allows for more principled, yet diverse learning rate schedules to be discovered.

EPBT can also be scaled up to larger DNN architectures with more weights. In Figure~\ref{fig:cifar10_wide}, Baseline 1 and EPBT were used to train four different wide residual networks \cite{zagoruyko2016wide} with different number of layers, but with similar number of parameters (11 million). The four architectures, in order of increasing depth, are: WRN-10-12, WRN-16-8, WRN-22-6, and WRN-28-5. EPBT is able to achieve noticeable improvements over Baseline 1 for all of the networks, reaching a better test accuracy at a faster pace. However, the final performance difference between the baseline and EPBT does decrease as the architecture becomes more complicated. Recent work has shown that DNNs are implicitly self-regularizing and thus might benefit less from external regularization as they increase in depth and size \cite{martin2018implicit}.

\subsection{SVHN}

To demonstrate that loss function optimization scales with dataset size, EPBT was applied to SVHN \cite{netzer2011reading}, a larger image classification task. This dataset is composed of around 600,000 training images and 26,000 testing images. Following existing practices \cite{huang2017densely}, the dataset was normalized but no data augmentation was used during training. The baseline model was optimized with SGD on the full training set for a total of 40 epochs, with the learning rate decaying from 0.1 by a factor of 10 at 20 and 30 epochs. EPBT was run for 40 generations, each with one epoch of training, and a validation set of 30,000 images was separated for evaluating individuals. Otherwise, the experiment setup was identical to the CIFAR-10 domain.

Figure~\ref{fig:svhn_res} gives a comparison of EPBT against Baseline 1 in the SVHN domain. As expected, EPBT achieves higher test accuracy than the baseline. Like in the earlier experiments with CIFAR-10 and ResNet-32, both EPBT variants learn faster and converge to a high test accuracy at the end. Interestingly, while the baseline begins to overfit and drop in accuracy at the end of training, EPBT's regularization mechanisms allow it to avoid this effect and maintain performance. 

\begin{table*}[t]
  \begin{center}
  \begin{footnotesize}
  \begin{sc}
  \begin{tabular}{lcccccc}
  \toprule
  Algorithm & CIFAR-10, ResNet-32 & CIFAR-10, WRN-10-12 & CIFAR-10, WRN-16-8 &  CIFAR-10, WRN-22-6 & CIFAR-10, WRN-28-5 & SVHN, ResNet-32 \\
  \midrule
  Baseline 1 (no PBT) & 168 & 168 & 168 & 176 & 184 & 21 \\ 
  Baseline 2 (PBT) & 128 & -- & -- & -- & -- & -- \\
  \bottomrule
  \end{tabular}
  \end{sc}
  \end{footnotesize}
  \end{center}
  \vskip 0.15in
  \caption{Number of training epochs required for EPBT to exceed the final test accuracy of the baselines. The baselines were trained for 200 epochs in CIFAR-10 and 40 epochs in SVHN. EPBT surpasses most of the baselines at around 80\% into the run.}
  \label{table:epbt_results2}
  \vspace*{-3ex}
\end{table*}

\section{Analysis of Results}

This section analyzes the performance and computational complexity of EPBT, as well as the loss functions and learning rate schedules that EPBT discovered.

\subsection{Performance}

A summary of the final test accuracies at the end of training for EPBT and the baselines is shown in Table~\ref{table:epbt_results}. The results show that EPBT achieves the best results for multiple datasets and model architectures. Another noticeable benefit provided by EPBT is the ability to train models to convergence significantly faster than non-population based methods, especially with a limited number of training epochs. This is because multiple models are simultaneously trained with EPBT, each with different loss functions. If progress is made in one of the models, its higher fitness leads to that model's loss function or weights being shared among the rest of the models, thus lifting their performance as well. 

Table~\ref{table:epbt_results2} details how many epochs of training are required for EPBT to surpass the fully trained performance of the baselines. As expected, EPBT outperforms the baselines on most architectures after training for 80\% the total number of epochs. The results are remarkable considering that Baseline 1 is trained on the full training set, while EPBT is not. These experiments thus demonstrate the power of EPBT in not just training better models but doing it faster too. 

The experiments also suggest that EPBT's main components, i.e.\ the metalearning evolutionary loop, genetic operators, Novelty Selection and Pulsation, TaylorGLO parameterization, and PBD, serve as powerful metalearning and regularization tools when combined. While each component has its own weaknesses, other components can help mitigate them. For example, TaylorGLO is useful for its regularizing effects, but the search space is large and potentially deceptive. However, Novelty Selection and Pulsation can help overcome this deception by maintaining population diversity. Similarly, PBD is a powerful general regularization tool that requires a good teacher model to work. Conveniently, such a model is provided by the elite individuals in EPBT's population.

\subsection{Computational Complexity}

Compared to simpler hyperparameter tuning methods that do not interleave training and optimization, EPBT is much more efficient. On the CIFAR-10 dataset, EPBT discovered 40 new loss functions during the first generation and an additional 20 loss functions every subsequent generation. EPBT was run for 25 generations and thus was able to explore up to 520 unique TaylorGLO parameterizations. This process is efficient despite how large the search space is; if grid search is performed at intervals of $1.0$, a total of $21^8$ (38 billion) unique loss function parameterizations will have to be evaluated. 

Furthermore, the computational complexity of EPBT scales linearly with the population size and not with the number of loss functions explored. Loss function evaluation is efficient in EPBT because it is not necessary to retrain the model from scratch whenever a new loss function is discovered; the model's weights are copied over from an existing model with good performance. If each of the 520 discovered loss functions was used to fully train a model from random initialization, over 100,000 epochs of training would be required, much higher than the 8,000 epochs EPBT needed. 

Because EPBT evaluates all the individuals in the population in parallel, the real-time complexity of each generation is not significantly higher than training a single model for the same number of epochs. Furthermore, the amount of time spent in Steps 1 and 2 to generate new individuals is negligible compared to Step 3, where model training occurs. The EPBT experiments in this paper were run on a machine with eight NVIDIA V100 GPUs and utilized several GPU-days worth of compute. When comparing computational cost to EPBT, Baseline 1 used 40 times less compute while Baseline 2 (PBT) required the same amount.

\begin{figure}[t]
  \centering
  \includegraphics[width=\linewidth]{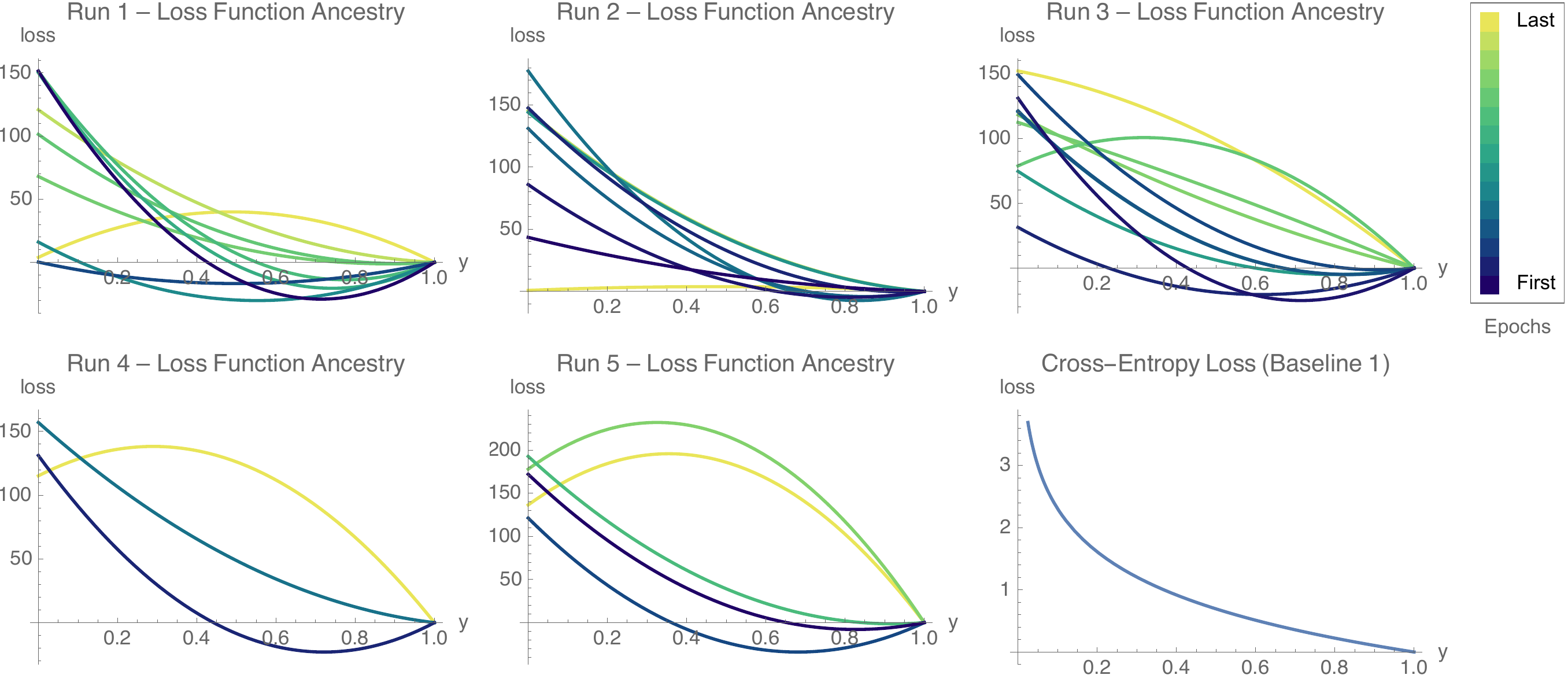}
  \caption{EPBT loss function ancestries for the best candidates across five different runs on CIFAR-10 with ResNet-32. Their shapes are simplified into a 2D binary classification loss \cite{gonzalez2019glo} for visualization purposes. Cross-entropy loss is shown in the bottom right plot for comparison. Loss functions in the starting generations with fewer training epochs are darker, while functions from later generations with more training are lighter. Some runs have fewer ancestors because elitism allows the same loss function to be reused for multiple generations. Across all runs, there is a temporal pattern in the loss function ancestry. Early loss functions tend to regularize more (indicated by a positive slope at $y=1$), while later functions encourage more accurate fitting to the ground-truth labels.}
  \label{fig:loss_change}
\end{figure}

\subsection{Loss Functions}

Do the loss functions discovered by EPBT stay static or adapt to the current stage of the training process? An analysis of functions discovered by EPBT during an experiment indicates that they do change significantly over the generations.

To characterize how the loss functions adapt with increased training, the ancestries for the final top-performing functions across five separate runs of EPBT (CIFAR-10, ResNet-32) are shown in Figure~\ref{fig:loss_change}. The cross-entropy loss (as used in Baseline 1) is plotted for comparison as well. Ancestry is determined by tracing the sequence of individuals $M_{0i} \ldots M_{ni}$, where $M_{(g-1)i}$ is the parent whose $\mathbf{D}_{(g-1)i}$ and $\mathbf{h}_{(g-1)i}$ were used to create $M_{gi}$. The sequence is simplified by removing any duplicate individuals that do not change between generations due to elitism, thus causing some runs to have shorter ancestries. 

Because the loss functions are multidimensional, graphing them is not straightforward. However, for visualization purposes, the losses can be simplified into a 2D binary classification modality where $y=1$ represents a perfect prediction, and $y=0$ represents a completely incorrect prediction \cite{gonzalez2019glo}. This approach makes it clear that the loss generally decreases as the predicted labels become more accurate and closer to the ground-truth labels. 

There is an interesting trend across all five runs: the loss functions optimized by EPBT are not all monotonically-decreasing. Instead, there are parabolic losses that have a minimum of around $0.7$ and rises slightly as $y$ approaches $1$. Such concavity is likely a form of regularization that prevents the network from overfitting to the training data by penalizing low-entropy prediction distributions centered around $y=1$. Similar behavior was observed when training using GLO \cite{gonzalez2019glo}.

 The plots also show that the loss functions change shape as training progresses. As the number of epochs increases, the slope near $y=1$ becomes increasingly positive, suggesting less regularization would occur. This result is consistent with recent research that suggests regularization is most important during a critical period early in the training process \cite{golatkar2019time}. If regularization is reduced or removed after this critical period, generalization sometimes may even improve. In EPBT, this principle was discovered and optimized without any prior knowledge as part of the metalearning process. EPBT thus provides an automatic way for exploring metaknowledge that could be difficult to come upon manually. 

Since different stages of EPBT utilize different types of loss functions, it is possible that a single static loss might not be optimal for the entire training process. Furthermore, loss functions that change makes sense considering that the learning dynamics for some DNNs are non-stationary or unstable \cite{jaderberg2017population}. For example, adaptive losses might improve the training of generative adversarial networks \cite{goodfellow2014generative, radford2015unsupervised}.

\begin{figure}[t]
  \centering
   \includegraphics[width=0.49\linewidth]{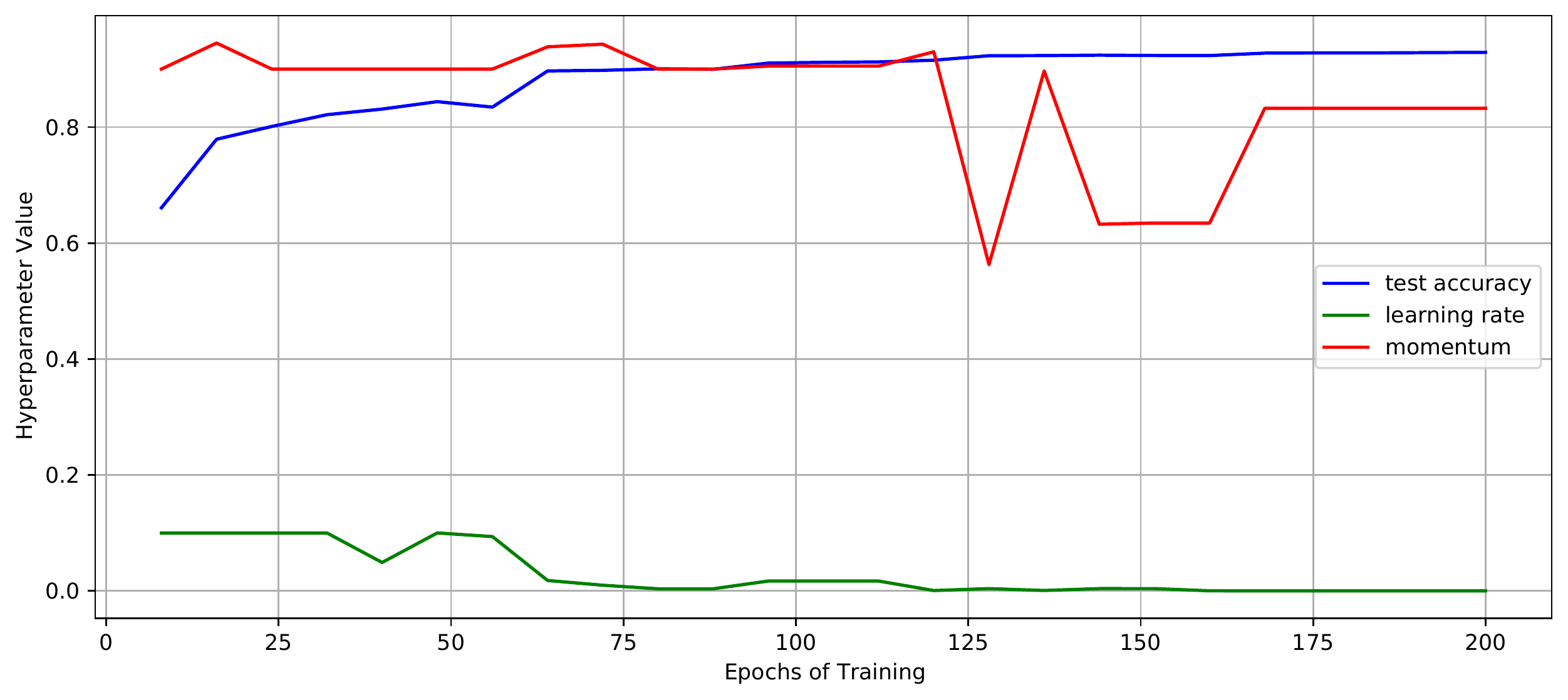}
   \includegraphics[width=0.49\linewidth]{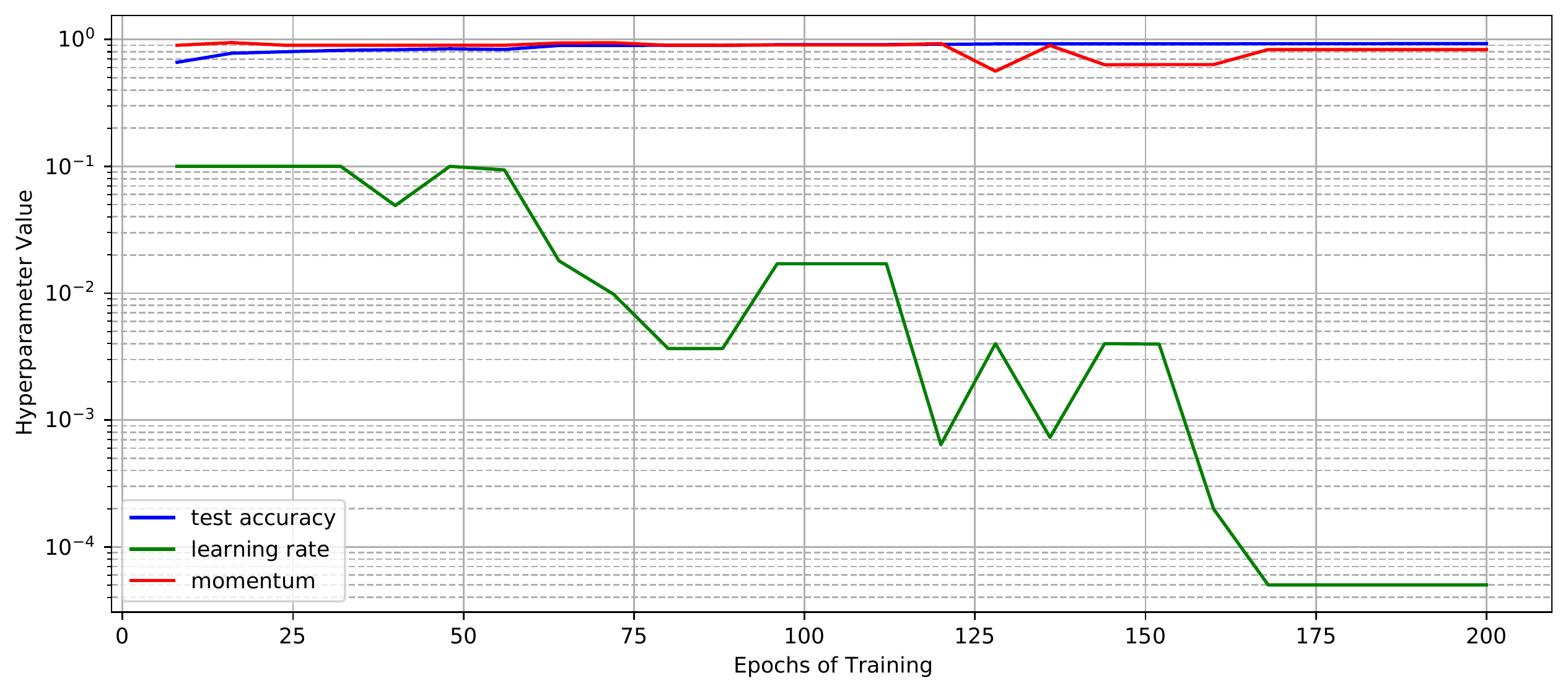}
  \caption{Visualization of how the learning rate and momentum of the best individual in the population changes during an EPBT run. The right plot is a log-scale version of the left plot. A cyclic and decaying learning rate pattern is present, while no such pattern exists for momentum.}
  \label{fig:sgd_hyperparam}
\end{figure}

\subsection{Learning Rate Schedules}

In addition to the TaylorGLO parameters, EPBT also optimized SGD hyperparameters such as learning rate schedule and momentum. Figure~\ref{fig:sgd_hyperparam} shows how these parameters of the best individual in the population changes over the course of an EPBT run (CIFAR-10, ResNet-32). While momentum remains mostly the same with some occasional dips and bumps, there is a clear downward, decaying trend for learning rate. The discovered learning rate schedule shares several similarities with hand designed schedules: both use high learning rates early in training for rapid learning but lower learning rates later to fine tune the model weights. More interestingly, there appears to be several cycles in the EPBT optimized schedule where the learning rate repeatedly goes up and down. This might be due the beneficial effects of cyclic learning rates in helping SGD escape from saddle points in the loss landscape during training \cite{smith2017cyclical}. 

\section{Discussion and Future Work}

When viewed from the EA perspective, EPBT is a more complex variant of PBT \cite{jaderberg2017population}. Mutation corresponds to the \textit{explore} step and elitism corresponds to the \textit{exploit} step in PBT. Besides Novelty Pulsation, EPBT improves upon PBT in two major ways. First, EPBT makes use of uniform Gaussian mutation (compared to the deterministic mutation in PBT) and uniform crossover. These biologically inspired heuristics allow PBT to scale better to higher dimensions. In particular, the crossover operator plays an important role in discovering good global solutions in large search spaces \cite{goldberg1988genetic, whitley1994genetic}. Second, EPBT utilizes tournament selection, a heuristic that helps prevent premature convergence to a local optimum \cite{shukla2015comparative}.

One possible direction of future work is be to allow the loss function parameterization to take the current state of training as input, potentially resulting in more refined training and better performance. Alternatively, domain information could be taken into account, allowing the learned loss function schedules to be more easily transferred to different tasks. EPBT could also be extended to include neural architecture search to jointly optimize network structure and loss functions. A more detailed study on how network architecture impacts EPBT performance and what components are necessary for training a particular DNN architectures could also lead to better models in the future. Another compelling direction would be to understand the synergies of the various components of EPBT. Ablation studies can be used to characterize the contribution that each component provides. Lastly, more experimental runs can be performed to improve statistical analysis of EPBT and the baselines.

\section{Conclusion}

This paper presents EPBT, an evolutionary algorithm for regularization metalearning. EPBT first improves upon PBT by introducing more advanced genetic operators. It then focuses it on regularization by evolving TaylorGLO loss functions. The deceptive interactions of weight and loss adaptation require more diversity, which is achieved through Novelty Pulsation, and more careful avoidance of overfitting, which is achieved through Population-Based Distillation.  On the CIFAR-10 and SVHN image classification benchmarks with several ResNet and Wide Resnet architectures, EPBT achieved faster and better model training. An analysis of the optimized loss functions suggests that these advantages stem from discovering strong regularization automatically. Furthermore, an adaptive loss function schedule naturally emerges as a likely key to achieving such performance. EPBT thus forms a practical method for regularization metalearning in deep networks.

\bibliographystyle{ACM-Reference-Format}
\bibliography{paper}


\begin{thebibliography}{58}


\ifx \showCODEN    \undefined \def \showCODEN     #1{\unskip}     \fi
\ifx \showDOI      \undefined \def \showDOI       #1{#1}\fi
\ifx \showISBNx    \undefined \def \showISBNx     #1{\unskip}     \fi
\ifx \showISBNxiii \undefined \def \showISBNxiii  #1{\unskip}     \fi
\ifx \showISSN     \undefined \def \showISSN      #1{\unskip}     \fi
\ifx \showLCCN     \undefined \def \showLCCN      #1{\unskip}     \fi
\ifx \shownote     \undefined \def \shownote      #1{#1}          \fi
\ifx \showarticletitle \undefined \def \showarticletitle #1{#1}   \fi
\ifx \showURL      \undefined \def \showURL       {\relax}        \fi
\providecommand\bibfield[2]{#2}
\providecommand\bibinfo[2]{#2}
\providecommand\natexlab[1]{#1}
\providecommand\showeprint[2][]{arXiv:#2}

\bibitem[\protect\citeauthoryear{Bahdanau, Cho, and Bengio}{Bahdanau
  et~al\mbox{.}}{2014}]%
        {bahdanau2014neural}
\bibfield{author}{\bibinfo{person}{Dzmitry Bahdanau},
  \bibinfo{person}{Kyunghyun Cho}, {and} \bibinfo{person}{Yoshua Bengio}.}
  \bibinfo{year}{2014}\natexlab{}.
\newblock \showarticletitle{Neural machine translation by jointly learning to
  align and translate}.
\newblock \bibinfo{journal}{\emph{arXiv preprint arXiv:1409.0473}}
  (\bibinfo{year}{2014}).
\newblock


\bibitem[\protect\citeauthoryear{Balaji, Sankaranarayanan, and
  Chellappa}{Balaji et~al\mbox{.}}{2018}]%
        {balaji2018metareg}
\bibfield{author}{\bibinfo{person}{Yogesh Balaji}, \bibinfo{person}{Swami
  Sankaranarayanan}, {and} \bibinfo{person}{Rama Chellappa}.}
  \bibinfo{year}{2018}\natexlab{}.
\newblock \showarticletitle{Metareg: Towards domain generalization using
  meta-regularization}.
\newblock \bibinfo{journal}{\emph{Advances in Neural Information Processing
  Systems}}  \bibinfo{volume}{31} (\bibinfo{year}{2018}),
  \bibinfo{pages}{998--1008}.
\newblock


\bibitem[\protect\citeauthoryear{Banzhaf, Nordin, Keller, and Francone}{Banzhaf
  et~al\mbox{.}}{1998}]%
        {banzhaf1998genetic}
\bibfield{author}{\bibinfo{person}{Wolfgang Banzhaf}, \bibinfo{person}{Peter
  Nordin}, \bibinfo{person}{Robert~E Keller}, {and} \bibinfo{person}{Frank~D
  Francone}.} \bibinfo{year}{1998}\natexlab{}.
\newblock \bibinfo{booktitle}{\emph{Genetic programming: an introduction}}.
  Vol.~\bibinfo{volume}{1}.
\newblock \bibinfo{publisher}{Morgan Kaufmann San Francisco}.
\newblock


\bibitem[\protect\citeauthoryear{Bosman, Engelbrecht, and Helbig}{Bosman
  et~al\mbox{.}}{2019}]%
        {bosman2019visualising}
\bibfield{author}{\bibinfo{person}{Anna~Sergeevna Bosman},
  \bibinfo{person}{Andries Engelbrecht}, {and} \bibinfo{person}{Mard{\'e}
  Helbig}.} \bibinfo{year}{2019}\natexlab{}.
\newblock \showarticletitle{Visualising Basins of Attraction for the
  Cross-Entropy and the Squared Error Neural Network Loss Functions}.
\newblock \bibinfo{journal}{\emph{arXiv preprint arXiv:1901.02302}}
  (\bibinfo{year}{2019}).
\newblock


\bibitem[\protect\citeauthoryear{Cubuk, Zoph, Mane, Vasudevan, and Le}{Cubuk
  et~al\mbox{.}}{2018}]%
        {cubuk2018autoaugment}
\bibfield{author}{\bibinfo{person}{Ekin~D Cubuk}, \bibinfo{person}{Barret
  Zoph}, \bibinfo{person}{Dandelion Mane}, \bibinfo{person}{Vijay Vasudevan},
  {and} \bibinfo{person}{Quoc~V Le}.} \bibinfo{year}{2018}\natexlab{}.
\newblock \showarticletitle{Autoaugment: Learning augmentation policies from
  data}.
\newblock \bibinfo{journal}{\emph{arXiv preprint arXiv:1805.09501}}
  (\bibinfo{year}{2018}).
\newblock


\bibitem[\protect\citeauthoryear{DeVries and Taylor}{DeVries and
  Taylor}{2017}]%
        {devries2017improved}
\bibfield{author}{\bibinfo{person}{Terrance DeVries} {and}
  \bibinfo{person}{Graham~W Taylor}.} \bibinfo{year}{2017}\natexlab{}.
\newblock \showarticletitle{Improved regularization of convolutional neural
  networks with cutout}.
\newblock \bibinfo{journal}{\emph{arXiv preprint arXiv:1708.04552}}
  (\bibinfo{year}{2017}).
\newblock


\bibitem[\protect\citeauthoryear{Golatkar, Achille, and Soatto}{Golatkar
  et~al\mbox{.}}{2019}]%
        {golatkar2019time}
\bibfield{author}{\bibinfo{person}{Aditya~Sharad Golatkar},
  \bibinfo{person}{Alessandro Achille}, {and} \bibinfo{person}{Stefano
  Soatto}.} \bibinfo{year}{2019}\natexlab{}.
\newblock \showarticletitle{Time Matters in Regularizing Deep Networks: Weight
  Decay and Data Augmentation Affect Early Learning Dynamics, Matter Little
  Near Convergence}. In \bibinfo{booktitle}{\emph{Advances in Neural
  Information Processing Systems 32}}. \bibinfo{pages}{10677--10687}.
\newblock


\bibitem[\protect\citeauthoryear{Goldberg and Holland}{Goldberg and
  Holland}{1988}]%
        {goldberg1988genetic}
\bibfield{author}{\bibinfo{person}{David~E Goldberg} {and}
  \bibinfo{person}{John~H Holland}.} \bibinfo{year}{1988}\natexlab{}.
\newblock \showarticletitle{Genetic algorithms and machine learning}.
\newblock \bibinfo{journal}{\emph{Machine learning}} \bibinfo{volume}{3},
  \bibinfo{number}{2} (\bibinfo{year}{1988}), \bibinfo{pages}{95--99}.
\newblock


\bibitem[\protect\citeauthoryear{Gonzalez and Miikkulainen}{Gonzalez and
  Miikkulainen}{2019}]%
        {gonzalez2019glo}
\bibfield{author}{\bibinfo{person}{Santiago Gonzalez} {and}
  \bibinfo{person}{Risto Miikkulainen}.} \bibinfo{year}{2019}\natexlab{}.
\newblock \showarticletitle{Improved Training Speed, Accuracy, and Data
  Utilization Through Loss Function Optimization}.
\newblock \bibinfo{journal}{\emph{arXiv preprint arXiv:1905.11528}}
  (\bibinfo{year}{2019}).
\newblock


\bibitem[\protect\citeauthoryear{Gonzalez and Miikkulainen}{Gonzalez and
  Miikkulainen}{2020}]%
        {gonzalez2020taylorglo}
\bibfield{author}{\bibinfo{person}{Santiago Gonzalez} {and}
  \bibinfo{person}{Risto Miikkulainen}.} \bibinfo{year}{2020}\natexlab{}.
\newblock \showarticletitle{Evolving Loss Functions With Multivariate {T}aylor
  Polynomial Parameterizations}.
\newblock \bibinfo{journal}{\emph{arXiv preprint arXiv:2002.00059}}
  (\bibinfo{year}{2020}).
\newblock


\bibitem[\protect\citeauthoryear{Goodfellow, Pouget-Abadie, Mirza, Xu,
  Warde-Farley, Ozair, Courville, and Bengio}{Goodfellow
  et~al\mbox{.}}{2014a}]%
        {goodfellow2014generative}
\bibfield{author}{\bibinfo{person}{Ian Goodfellow}, \bibinfo{person}{Jean
  Pouget-Abadie}, \bibinfo{person}{Mehdi Mirza}, \bibinfo{person}{Bing Xu},
  \bibinfo{person}{David Warde-Farley}, \bibinfo{person}{Sherjil Ozair},
  \bibinfo{person}{Aaron Courville}, {and} \bibinfo{person}{Yoshua Bengio}.}
  \bibinfo{year}{2014}\natexlab{a}.
\newblock \showarticletitle{Generative adversarial nets}. In
  \bibinfo{booktitle}{\emph{Advances in Neural Information Processing
  Systems}}. \bibinfo{pages}{2672--2680}.
\newblock


\bibitem[\protect\citeauthoryear{Goodfellow, Shlens, and Szegedy}{Goodfellow
  et~al\mbox{.}}{2014b}]%
        {goodfellow2014explaining}
\bibfield{author}{\bibinfo{person}{Ian~J Goodfellow}, \bibinfo{person}{Jonathon
  Shlens}, {and} \bibinfo{person}{Christian Szegedy}.}
  \bibinfo{year}{2014}\natexlab{b}.
\newblock \showarticletitle{Explaining and harnessing adversarial examples}.
\newblock \bibinfo{journal}{\emph{arXiv preprint arXiv:1412.6572}}
  (\bibinfo{year}{2014}).
\newblock


\bibitem[\protect\citeauthoryear{Hansen and Ostermeier}{Hansen and
  Ostermeier}{1996}]%
        {hansen1996cmaes}
\bibfield{author}{\bibinfo{person}{Nikolaus Hansen} {and}
  \bibinfo{person}{Andreas Ostermeier}.} \bibinfo{year}{1996}\natexlab{}.
\newblock \showarticletitle{Adapting arbitrary normal mutation distributions in
  evolution strategies: The covariance matrix adaptation}. In
  \bibinfo{booktitle}{\emph{Proceedings of IEEE international conference on
  evolutionary computation}}. IEEE, \bibinfo{pages}{312--317}.
\newblock


\bibitem[\protect\citeauthoryear{He, Zhang, Ren, and Sun}{He
  et~al\mbox{.}}{2015}]%
        {he2015delving}
\bibfield{author}{\bibinfo{person}{Kaiming He}, \bibinfo{person}{Xiangyu
  Zhang}, \bibinfo{person}{Shaoqing Ren}, {and} \bibinfo{person}{Jian Sun}.}
  \bibinfo{year}{2015}\natexlab{}.
\newblock \showarticletitle{Delving deep into rectifiers: Surpassing
  human-level performance on imagenet classification}. In
  \bibinfo{booktitle}{\emph{Proceedings of the IEEE international conference on
  computer vision}}. \bibinfo{pages}{1026--1034}.
\newblock


\bibitem[\protect\citeauthoryear{He, Zhang, Ren, and Sun}{He
  et~al\mbox{.}}{2016}]%
        {resnet}
\bibfield{author}{\bibinfo{person}{Kaiming He}, \bibinfo{person}{Xiangyu
  Zhang}, \bibinfo{person}{Shaoqing Ren}, {and} \bibinfo{person}{Jian Sun}.}
  \bibinfo{year}{2016}\natexlab{}.
\newblock \showarticletitle{Deep Residual Learning for Image Recognition}.
\newblock \bibinfo{journal}{\emph{IEEE Conference on Computer Vision and
  Pattern Recognition (CVPR)}} (\bibinfo{year}{2016}),
  \bibinfo{pages}{770--778}.
\newblock


\bibitem[\protect\citeauthoryear{Hinton, Vinyals, and Dean}{Hinton
  et~al\mbox{.}}{2015}]%
        {hinton2015distilling}
\bibfield{author}{\bibinfo{person}{Geoffrey Hinton}, \bibinfo{person}{Oriol
  Vinyals}, {and} \bibinfo{person}{Jeff Dean}.}
  \bibinfo{year}{2015}\natexlab{}.
\newblock \showarticletitle{Distilling the knowledge in a neural network}.
\newblock \bibinfo{journal}{\emph{arXiv preprint arXiv:1503.02531}}
  (\bibinfo{year}{2015}).
\newblock


\bibitem[\protect\citeauthoryear{Ho, Liang, Stoica, Abbeel, and Chen}{Ho
  et~al\mbox{.}}{2019}]%
        {ho2019population}
\bibfield{author}{\bibinfo{person}{Daniel Ho}, \bibinfo{person}{Eric Liang},
  \bibinfo{person}{Ion Stoica}, \bibinfo{person}{Pieter Abbeel}, {and}
  \bibinfo{person}{Xi Chen}.} \bibinfo{year}{2019}\natexlab{}.
\newblock \showarticletitle{Population based augmentation: Efficient learning
  of augmentation policy schedules}.
\newblock \bibinfo{journal}{\emph{arXiv preprint arXiv:1905.05393}}
  (\bibinfo{year}{2019}).
\newblock


\bibitem[\protect\citeauthoryear{Houthooft, Chen, Isola, Stadie, Wolski, Ho,
  and Abbeel}{Houthooft et~al\mbox{.}}{2018}]%
        {houthooft2018evolved}
\bibfield{author}{\bibinfo{person}{Rein Houthooft}, \bibinfo{person}{Yuhua
  Chen}, \bibinfo{person}{Phillip Isola}, \bibinfo{person}{Bradly Stadie},
  \bibinfo{person}{Filip Wolski}, \bibinfo{person}{OpenAI~Jonathan Ho}, {and}
  \bibinfo{person}{Pieter Abbeel}.} \bibinfo{year}{2018}\natexlab{}.
\newblock \showarticletitle{Evolved policy gradients}. In
  \bibinfo{booktitle}{\emph{Advances in Neural Information Processing
  Systems}}. \bibinfo{pages}{5400--5409}.
\newblock


\bibitem[\protect\citeauthoryear{Huang, Liu, Van Der~Maaten, and
  Weinberger}{Huang et~al\mbox{.}}{2017}]%
        {huang2017densely}
\bibfield{author}{\bibinfo{person}{Gao Huang}, \bibinfo{person}{Zhuang Liu},
  \bibinfo{person}{Laurens Van Der~Maaten}, {and} \bibinfo{person}{Kilian~Q
  Weinberger}.} \bibinfo{year}{2017}\natexlab{}.
\newblock \showarticletitle{Densely connected convolutional networks}. In
  \bibinfo{booktitle}{\emph{Proceedings of the IEEE conference on computer
  vision and pattern recognition}}. \bibinfo{pages}{4700--4708}.
\newblock


\bibitem[\protect\citeauthoryear{Ioffe and Szegedy}{Ioffe and Szegedy}{2015}]%
        {ioffe2015batch}
\bibfield{author}{\bibinfo{person}{Sergey Ioffe} {and}
  \bibinfo{person}{Christian Szegedy}.} \bibinfo{year}{2015}\natexlab{}.
\newblock \showarticletitle{Batch normalization: Accelerating deep network
  training by reducing internal covariate shift}. In
  \bibinfo{booktitle}{\emph{International conference on machine learning}}.
  PMLR, \bibinfo{pages}{448--456}.
\newblock


\bibitem[\protect\citeauthoryear{Jaderberg, Dalibard, Osindero, Czarnecki,
  Donahue, Razavi, Vinyals, Green, Dunning, Simonyan, et~al\mbox{.}}{Jaderberg
  et~al\mbox{.}}{2017}]%
        {jaderberg2017population}
\bibfield{author}{\bibinfo{person}{Max Jaderberg}, \bibinfo{person}{Valentin
  Dalibard}, \bibinfo{person}{Simon Osindero}, \bibinfo{person}{Wojciech~M
  Czarnecki}, \bibinfo{person}{Jeff Donahue}, \bibinfo{person}{Ali Razavi},
  \bibinfo{person}{Oriol Vinyals}, \bibinfo{person}{Tim Green},
  \bibinfo{person}{Iain Dunning}, \bibinfo{person}{Karen Simonyan},
  {et~al\mbox{.}}} \bibinfo{year}{2017}\natexlab{}.
\newblock \showarticletitle{Population based training of neural networks}.
\newblock \bibinfo{journal}{\emph{arXiv preprint arXiv:1711.09846}}
  (\bibinfo{year}{2017}).
\newblock


\bibitem[\protect\citeauthoryear{Janocha and Czarnecki}{Janocha and
  Czarnecki}{2017}]%
        {janocha2017loss}
\bibfield{author}{\bibinfo{person}{Katarzyna Janocha} {and}
  \bibinfo{person}{Wojciech~Marian Czarnecki}.}
  \bibinfo{year}{2017}\natexlab{}.
\newblock \showarticletitle{On loss functions for deep neural networks in
  classification}.
\newblock \bibinfo{journal}{\emph{arXiv preprint arXiv:1702.05659}}
  (\bibinfo{year}{2017}).
\newblock


\bibitem[\protect\citeauthoryear{Kim, Ji, Yoon, and Hwang}{Kim
  et~al\mbox{.}}{2020}]%
        {kim2020self}
\bibfield{author}{\bibinfo{person}{Kyungyul Kim}, \bibinfo{person}{ByeongMoon
  Ji}, \bibinfo{person}{Doyoung Yoon}, {and} \bibinfo{person}{Sangheum Hwang}.}
  \bibinfo{year}{2020}\natexlab{}.
\newblock \showarticletitle{Self-knowledge distillation: A simple way for
  better generalization}.
\newblock \bibinfo{journal}{\emph{arXiv preprint arXiv:2006.12000}}
  (\bibinfo{year}{2020}).
\newblock


\bibitem[\protect\citeauthoryear{Klein, Falkner, Bartels, Hennig, and
  Hutter}{Klein et~al\mbox{.}}{2017}]%
        {klein2017fast}
\bibfield{author}{\bibinfo{person}{Aaron Klein}, \bibinfo{person}{Stefan
  Falkner}, \bibinfo{person}{Simon Bartels}, \bibinfo{person}{Philipp Hennig},
  {and} \bibinfo{person}{Frank Hutter}.} \bibinfo{year}{2017}\natexlab{}.
\newblock \showarticletitle{Fast bayesian optimization of machine learning
  hyperparameters on large datasets}. In \bibinfo{booktitle}{\emph{Artificial
  Intelligence and Statistics}}. PMLR, \bibinfo{pages}{528--536}.
\newblock


\bibitem[\protect\citeauthoryear{Krizhevsky and Hinton}{Krizhevsky and
  Hinton}{2009}]%
        {krizhevsky2009learning}
\bibfield{author}{\bibinfo{person}{Alex Krizhevsky} {and}
  \bibinfo{person}{Geoffrey Hinton}.} \bibinfo{year}{2009}\natexlab{}.
\newblock \bibinfo{title}{Learning multiple layers of features from tiny
  images}.
\newblock
\newblock


\bibitem[\protect\citeauthoryear{Krizhevsky, Sutskever, and Hinton}{Krizhevsky
  et~al\mbox{.}}{2012}]%
        {NIPS2012_4824}
\bibfield{author}{\bibinfo{person}{Alex Krizhevsky}, \bibinfo{person}{Ilya
  Sutskever}, {and} \bibinfo{person}{Geoffrey~E Hinton}.}
  \bibinfo{year}{2012}\natexlab{}.
\newblock \showarticletitle{{ImageNet} Classification with Deep Convolutional
  Neural Networks}.
\newblock In \bibinfo{booktitle}{\emph{Advances in Neural Information
  Processing Systems 25}}, \bibfield{editor}{\bibinfo{person}{F.~Pereira},
  \bibinfo{person}{C.~J.~C. Burges}, \bibinfo{person}{L.~Bottou}, {and}
  \bibinfo{person}{K.~Q. Weinberger}} (Eds.). \bibinfo{publisher}{Curran
  Associates, Inc.}, \bibinfo{pages}{1097--1105}.
\newblock


\bibitem[\protect\citeauthoryear{Kuka{\v{c}}ka, Golkov, and
  Cremers}{Kuka{\v{c}}ka et~al\mbox{.}}{2017}]%
        {kukavcka2017regularization}
\bibfield{author}{\bibinfo{person}{Jan Kuka{\v{c}}ka},
  \bibinfo{person}{Vladimir Golkov}, {and} \bibinfo{person}{Daniel Cremers}.}
  \bibinfo{year}{2017}\natexlab{}.
\newblock \showarticletitle{Regularization for deep learning: A taxonomy}.
\newblock \bibinfo{journal}{\emph{arXiv preprint arXiv:1710.10686}}
  (\bibinfo{year}{2017}).
\newblock


\bibitem[\protect\citeauthoryear{LeCun, Bengio, and Hinton}{LeCun
  et~al\mbox{.}}{2015}]%
        {lecun2015deep}
\bibfield{author}{\bibinfo{person}{Yann LeCun}, \bibinfo{person}{Yoshua
  Bengio}, {and} \bibinfo{person}{Geoffrey Hinton}.}
  \bibinfo{year}{2015}\natexlab{}.
\newblock \showarticletitle{Deep learning}.
\newblock \bibinfo{journal}{\emph{Nature}} \bibinfo{volume}{521},
  \bibinfo{number}{7553} (\bibinfo{year}{2015}), \bibinfo{pages}{436}.
\newblock


\bibitem[\protect\citeauthoryear{Lehman and Stanley}{Lehman and
  Stanley}{2008}]%
        {lehman2008exploiting}
\bibfield{author}{\bibinfo{person}{Joel Lehman} {and}
  \bibinfo{person}{Kenneth~O Stanley}.} \bibinfo{year}{2008}\natexlab{}.
\newblock \showarticletitle{Exploiting open-endedness to solve problems through
  the search for novelty.}. In \bibinfo{booktitle}{\emph{ALIFE}}. Citeseer,
  \bibinfo{pages}{329--336}.
\newblock


\bibitem[\protect\citeauthoryear{Li, Jamieson, DeSalvo, Rostamizadeh, and
  Talwalkar}{Li et~al\mbox{.}}{2017}]%
        {li2017hyperband}
\bibfield{author}{\bibinfo{person}{Lisha Li}, \bibinfo{person}{Kevin Jamieson},
  \bibinfo{person}{Giulia DeSalvo}, \bibinfo{person}{Afshin Rostamizadeh},
  {and} \bibinfo{person}{Ameet Talwalkar}.} \bibinfo{year}{2017}\natexlab{}.
\newblock \showarticletitle{Hyperband: A novel bandit-based approach to
  hyperparameter optimization}.
\newblock \bibinfo{journal}{\emph{The Journal of Machine Learning Research}}
  \bibinfo{volume}{18}, \bibinfo{number}{1} (\bibinfo{year}{2017}),
  \bibinfo{pages}{6765--6816}.
\newblock


\bibitem[\protect\citeauthoryear{Liu, Simonyan, and Yang}{Liu
  et~al\mbox{.}}{2018}]%
        {liu2018darts}
\bibfield{author}{\bibinfo{person}{Hanxiao Liu}, \bibinfo{person}{Karen
  Simonyan}, {and} \bibinfo{person}{Yiming Yang}.}
  \bibinfo{year}{2018}\natexlab{}.
\newblock \showarticletitle{Darts: Differentiable architecture search}.
\newblock \bibinfo{journal}{\emph{arXiv preprint arXiv:1806.09055}}
  (\bibinfo{year}{2018}).
\newblock


\bibitem[\protect\citeauthoryear{Loshchilov and Hutter}{Loshchilov and
  Hutter}{2016}]%
        {loshchilov2016cma}
\bibfield{author}{\bibinfo{person}{Ilya Loshchilov} {and}
  \bibinfo{person}{Frank Hutter}.} \bibinfo{year}{2016}\natexlab{}.
\newblock \showarticletitle{{CMA-ES} for hyperparameter optimization of deep
  neural networks}.
\newblock \bibinfo{journal}{\emph{arXiv preprint arXiv:1604.07269}}
  (\bibinfo{year}{2016}).
\newblock


\bibitem[\protect\citeauthoryear{Lu, Whalen, Boddeti, Dhebar, Deb, Goodman, and
  Banzhaf}{Lu et~al\mbox{.}}{2018}]%
        {lu2018nsga}
\bibfield{author}{\bibinfo{person}{Zhichao Lu}, \bibinfo{person}{Ian Whalen},
  \bibinfo{person}{Vishnu Boddeti}, \bibinfo{person}{Yashesh Dhebar},
  \bibinfo{person}{Kalyanmoy Deb}, \bibinfo{person}{Erik Goodman}, {and}
  \bibinfo{person}{Wolfgang Banzhaf}.} \bibinfo{year}{2018}\natexlab{}.
\newblock \showarticletitle{NSGA-NET: a multi-objective genetic algorithm for
  neural architecture search}.
\newblock \bibinfo{journal}{\emph{arXiv preprint arXiv:1810.03522}}
  (\bibinfo{year}{2018}).
\newblock


\bibitem[\protect\citeauthoryear{Maclaurin, Duvenaud, and Adams}{Maclaurin
  et~al\mbox{.}}{2015}]%
        {maclaurin2015gradient}
\bibfield{author}{\bibinfo{person}{Dougal Maclaurin}, \bibinfo{person}{David
  Duvenaud}, {and} \bibinfo{person}{Ryan Adams}.}
  \bibinfo{year}{2015}\natexlab{}.
\newblock \showarticletitle{Gradient-based hyperparameter optimization through
  reversible learning}. In \bibinfo{booktitle}{\emph{International Conference
  on Machine Learning}}. \bibinfo{pages}{2113--2122}.
\newblock


\bibitem[\protect\citeauthoryear{Martin and Mahoney}{Martin and
  Mahoney}{2018}]%
        {martin2018implicit}
\bibfield{author}{\bibinfo{person}{Charles~H Martin} {and}
  \bibinfo{person}{Michael~W Mahoney}.} \bibinfo{year}{2018}\natexlab{}.
\newblock \showarticletitle{Implicit self-regularization in deep neural
  networks: Evidence from random matrix theory and implications for learning}.
\newblock \bibinfo{journal}{\emph{arXiv preprint arXiv:1810.01075}}
  (\bibinfo{year}{2018}).
\newblock


\bibitem[\protect\citeauthoryear{Miikkulainen, Liang, Meyerson, Rawal, Fink,
  Francon, Raju, Shahrzad, Navruzyan, Duffy, et~al\mbox{.}}{Miikkulainen
  et~al\mbox{.}}{2019}]%
        {miikkulainen2019evolving}
\bibfield{author}{\bibinfo{person}{Risto Miikkulainen}, \bibinfo{person}{Jason
  Liang}, \bibinfo{person}{Elliot Meyerson}, \bibinfo{person}{Aditya Rawal},
  \bibinfo{person}{Daniel Fink}, \bibinfo{person}{Olivier Francon},
  \bibinfo{person}{Bala Raju}, \bibinfo{person}{Hormoz Shahrzad},
  \bibinfo{person}{Arshak Navruzyan}, \bibinfo{person}{Nigel Duffy},
  {et~al\mbox{.}}} \bibinfo{year}{2019}\natexlab{}.
\newblock \showarticletitle{Evolving deep neural networks}.
\newblock In \bibinfo{booktitle}{\emph{Artificial Intelligence in the Age of
  Neural Networks and Brain Computing}}. \bibinfo{publisher}{Elsevier},
  \bibinfo{pages}{293--312}.
\newblock


\bibitem[\protect\citeauthoryear{Mnih, Kavukcuoglu, Silver, Rusu, Veness,
  Bellemare, Graves, Riedmiller, Fidjeland, Ostrovski, et~al\mbox{.}}{Mnih
  et~al\mbox{.}}{2015}]%
        {mnih2015human}
\bibfield{author}{\bibinfo{person}{Volodymyr Mnih}, \bibinfo{person}{Koray
  Kavukcuoglu}, \bibinfo{person}{David Silver}, \bibinfo{person}{Andrei~A
  Rusu}, \bibinfo{person}{Joel Veness}, \bibinfo{person}{Marc~G Bellemare},
  \bibinfo{person}{Alex Graves}, \bibinfo{person}{Martin Riedmiller},
  \bibinfo{person}{Andreas~K Fidjeland}, \bibinfo{person}{Georg Ostrovski},
  {et~al\mbox{.}}} \bibinfo{year}{2015}\natexlab{}.
\newblock \showarticletitle{Human-level control through deep reinforcement
  learning}.
\newblock \bibinfo{journal}{\emph{Nature}} \bibinfo{volume}{518},
  \bibinfo{number}{7540} (\bibinfo{year}{2015}), \bibinfo{pages}{529}.
\newblock


\bibitem[\protect\citeauthoryear{Moody, Hanson, Krogh, and Hertz}{Moody
  et~al\mbox{.}}{1995}]%
        {moody1995simple}
\bibfield{author}{\bibinfo{person}{John Moody}, \bibinfo{person}{Stephen
  Hanson}, \bibinfo{person}{Anders Krogh}, {and} \bibinfo{person}{John~A
  Hertz}.} \bibinfo{year}{1995}\natexlab{}.
\newblock \showarticletitle{A simple weight decay can improve generalization}.
\newblock \bibinfo{journal}{\emph{Advances in neural information processing
  systems}} \bibinfo{volume}{4}, \bibinfo{number}{1995} (\bibinfo{year}{1995}),
  \bibinfo{pages}{950--957}.
\newblock


\bibitem[\protect\citeauthoryear{Mouret and Clune}{Mouret and Clune}{2015}]%
        {mouret2015illuminating}
\bibfield{author}{\bibinfo{person}{Jean-Baptiste Mouret} {and}
  \bibinfo{person}{Jeff Clune}.} \bibinfo{year}{2015}\natexlab{}.
\newblock \showarticletitle{Illuminating search spaces by mapping elites}.
\newblock \bibinfo{journal}{\emph{arXiv preprint arXiv:1504.04909}}
  (\bibinfo{year}{2015}).
\newblock


\bibitem[\protect\citeauthoryear{M{\"u}ller, Kornblith, and Hinton}{M{\"u}ller
  et~al\mbox{.}}{2019}]%
        {muller2019does}
\bibfield{author}{\bibinfo{person}{Rafael M{\"u}ller}, \bibinfo{person}{Simon
  Kornblith}, {and} \bibinfo{person}{Geoffrey Hinton}.}
  \bibinfo{year}{2019}\natexlab{}.
\newblock \showarticletitle{When does label smoothing help?}
\newblock \bibinfo{journal}{\emph{arXiv preprint arXiv:1906.02629}}
  (\bibinfo{year}{2019}).
\newblock


\bibitem[\protect\citeauthoryear{Netzer, Wang, Coates, Bissacco, Wu, and
  Ng}{Netzer et~al\mbox{.}}{2011}]%
        {netzer2011reading}
\bibfield{author}{\bibinfo{person}{Yuval Netzer}, \bibinfo{person}{Tao Wang},
  \bibinfo{person}{Adam Coates}, \bibinfo{person}{Alessandro Bissacco},
  \bibinfo{person}{Bo Wu}, {and} \bibinfo{person}{Andrew~Y Ng}.}
  \bibinfo{year}{2011}\natexlab{}.
\newblock \bibinfo{title}{Reading digits in natural images with unsupervised
  feature learning}.
\newblock
\newblock


\bibitem[\protect\citeauthoryear{Pham, Guan, Zoph, Le, and Dean}{Pham
  et~al\mbox{.}}{2018}]%
        {pham2018efficient}
\bibfield{author}{\bibinfo{person}{Hieu Pham}, \bibinfo{person}{Melody~Y Guan},
  \bibinfo{person}{Barret Zoph}, \bibinfo{person}{Quoc~V Le}, {and}
  \bibinfo{person}{Jeff Dean}.} \bibinfo{year}{2018}\natexlab{}.
\newblock \showarticletitle{Efficient neural architecture search via parameter
  sharing}.
\newblock \bibinfo{journal}{\emph{arXiv preprint arXiv:1802.03268}}
  (\bibinfo{year}{2018}).
\newblock


\bibitem[\protect\citeauthoryear{Radford, Metz, and Chintala}{Radford
  et~al\mbox{.}}{2015}]%
        {radford2015unsupervised}
\bibfield{author}{\bibinfo{person}{Alec Radford}, \bibinfo{person}{Luke Metz},
  {and} \bibinfo{person}{Soumith Chintala}.} \bibinfo{year}{2015}\natexlab{}.
\newblock \showarticletitle{Unsupervised representation learning with deep
  convolutional generative adversarial networks}.
\newblock \bibinfo{journal}{\emph{arXiv preprint arXiv:1511.06434}}
  (\bibinfo{year}{2015}).
\newblock


\bibitem[\protect\citeauthoryear{Real, Aggarwal, Huang, and Le}{Real
  et~al\mbox{.}}{2019}]%
        {real2019regularized}
\bibfield{author}{\bibinfo{person}{Esteban Real}, \bibinfo{person}{Alok
  Aggarwal}, \bibinfo{person}{Yanping Huang}, {and} \bibinfo{person}{Quoc~V
  Le}.} \bibinfo{year}{2019}\natexlab{}.
\newblock \showarticletitle{Regularized evolution for image classifier
  architecture search}. In \bibinfo{booktitle}{\emph{Proceedings of the AAAI
  Conference on Artificial Intelligence}}, Vol.~\bibinfo{volume}{33}.
  \bibinfo{pages}{4780--4789}.
\newblock


\bibitem[\protect\citeauthoryear{Shahrzad, Fink, and Miikkulainen}{Shahrzad
  et~al\mbox{.}}{2018}]%
        {shahrzad2018noveltyselection}
\bibfield{author}{\bibinfo{person}{Hormoz Shahrzad}, \bibinfo{person}{Daniel
  Fink}, {and} \bibinfo{person}{Risto Miikkulainen}.}
  \bibinfo{year}{2018}\natexlab{}.
\newblock \showarticletitle{Enhanced Optimization with Composite Objectives and
  Novelty Selection}. \bibinfo{pages}{616--622}.
\newblock
\urldef\tempurl%
\url{https://doi.org/10.1162/isal_a_00113}
\showDOI{\tempurl}


\bibitem[\protect\citeauthoryear{Shahrzad, Hodjat, Dollé, Denissov, Lau,
  Goodhew, Dyer, and Miikkulainen}{Shahrzad et~al\mbox{.}}{2020}]%
        {shahrzad2020noveltypulsation}
\bibfield{author}{\bibinfo{person}{Hormoz Shahrzad}, \bibinfo{person}{Babak
  Hodjat}, \bibinfo{person}{Camille Dollé}, \bibinfo{person}{Andrei Denissov},
  \bibinfo{person}{Simon Lau}, \bibinfo{person}{Donn Goodhew},
  \bibinfo{person}{Justin Dyer}, {and} \bibinfo{person}{Risto Miikkulainen}.}
  \bibinfo{year}{2020}\natexlab{}.
\newblock \bibinfo{booktitle}{\emph{Enhanced Optimization with Composite
  Objectives and Novelty Pulsation}}.
\newblock \bibinfo{pages}{275--293}.
\newblock
\showISBNx{978-3-030-39957-3}
\urldef\tempurl%
\url{https://doi.org/10.1007/978-3-030-39958-0_14}
\showDOI{\tempurl}


\bibitem[\protect\citeauthoryear{Shukla, Pandey, and Mehrotra}{Shukla
  et~al\mbox{.}}{2015}]%
        {shukla2015comparative}
\bibfield{author}{\bibinfo{person}{Anupriya Shukla},
  \bibinfo{person}{Hari~Mohan Pandey}, {and} \bibinfo{person}{Deepti
  Mehrotra}.} \bibinfo{year}{2015}\natexlab{}.
\newblock \showarticletitle{Comparative review of selection techniques in
  genetic algorithm}. In \bibinfo{booktitle}{\emph{2015 International
  Conference on Futuristic Trends on Computational Analysis and Knowledge
  Management (ABLAZE)}}. IEEE, \bibinfo{pages}{515--519}.
\newblock


\bibitem[\protect\citeauthoryear{Smith}{Smith}{2017}]%
        {smith2017cyclical}
\bibfield{author}{\bibinfo{person}{Leslie~N Smith}.}
  \bibinfo{year}{2017}\natexlab{}.
\newblock \showarticletitle{Cyclical learning rates for training neural
  networks}. In \bibinfo{booktitle}{\emph{2017 IEEE winter conference on
  applications of computer vision (WACV)}}. IEEE, \bibinfo{pages}{464--472}.
\newblock


\bibitem[\protect\citeauthoryear{Snoek, Larochelle, and Adams}{Snoek
  et~al\mbox{.}}{2012}]%
        {snoek2012practical}
\bibfield{author}{\bibinfo{person}{Jasper Snoek}, \bibinfo{person}{Hugo
  Larochelle}, {and} \bibinfo{person}{Ryan~P Adams}.}
  \bibinfo{year}{2012}\natexlab{}.
\newblock \showarticletitle{Practical bayesian optimization of machine learning
  algorithms}. In \bibinfo{booktitle}{\emph{Advances in neural information
  processing systems}}. \bibinfo{pages}{2951--2959}.
\newblock


\bibitem[\protect\citeauthoryear{Snoek, Rippel, Swersky, Kiros, Satish,
  Sundaram, Patwary, Prabhat, and Adams}{Snoek et~al\mbox{.}}{2015}]%
        {snoek2015scalable}
\bibfield{author}{\bibinfo{person}{Jasper Snoek}, \bibinfo{person}{Oren
  Rippel}, \bibinfo{person}{Kevin Swersky}, \bibinfo{person}{Ryan Kiros},
  \bibinfo{person}{Nadathur Satish}, \bibinfo{person}{Narayanan Sundaram},
  \bibinfo{person}{Mostofa Patwary}, \bibinfo{person}{Mr Prabhat}, {and}
  \bibinfo{person}{Ryan Adams}.} \bibinfo{year}{2015}\natexlab{}.
\newblock \showarticletitle{Scalable bayesian optimization using deep neural
  networks}. In \bibinfo{booktitle}{\emph{International conference on machine
  learning}}. \bibinfo{pages}{2171--2180}.
\newblock


\bibitem[\protect\citeauthoryear{Srivastava, Hinton, Krizhevsky, Sutskever, and
  Salakhutdinov}{Srivastava et~al\mbox{.}}{2014}]%
        {srivastava2014dropout}
\bibfield{author}{\bibinfo{person}{Nitish Srivastava},
  \bibinfo{person}{Geoffrey Hinton}, \bibinfo{person}{Alex Krizhevsky},
  \bibinfo{person}{Ilya Sutskever}, {and} \bibinfo{person}{Ruslan
  Salakhutdinov}.} \bibinfo{year}{2014}\natexlab{}.
\newblock \showarticletitle{Dropout: a simple way to prevent neural networks
  from overfitting}.
\newblock \bibinfo{journal}{\emph{The journal of machine learning research}}
  \bibinfo{volume}{15}, \bibinfo{number}{1} (\bibinfo{year}{2014}),
  \bibinfo{pages}{1929--1958}.
\newblock


\bibitem[\protect\citeauthoryear{Stanley, Clune, Lehman, and
  Miikkulainen}{Stanley et~al\mbox{.}}{2019}]%
        {nature_neuroevolution}
\bibfield{author}{\bibinfo{person}{Kenneth~O. Stanley}, \bibinfo{person}{Jeff
  Clune}, \bibinfo{person}{Joel Lehman}, {and} \bibinfo{person}{Risto
  Miikkulainen}.} \bibinfo{year}{2019}\natexlab{}.
\newblock \showarticletitle{Designing neural networks through neuroevolution}.
\newblock \bibinfo{journal}{\emph{Nature Machine Intelligence}}
  \bibinfo{volume}{1}, \bibinfo{number}{1} (\bibinfo{year}{2019}),
  \bibinfo{pages}{24--35}.
\newblock
\showISSN{2522-5839}
\urldef\tempurl%
\url{https://doi.org/10.1038/s42256-018-0006-z}
\showDOI{\tempurl}


\bibitem[\protect\citeauthoryear{Whitley}{Whitley}{1994}]%
        {whitley1994genetic}
\bibfield{author}{\bibinfo{person}{Darrell Whitley}.}
  \bibinfo{year}{1994}\natexlab{}.
\newblock \showarticletitle{A genetic algorithm tutorial}.
\newblock \bibinfo{journal}{\emph{Statistics and computing}}
  \bibinfo{volume}{4}, \bibinfo{number}{2} (\bibinfo{year}{1994}),
  \bibinfo{pages}{65--85}.
\newblock


\bibitem[\protect\citeauthoryear{You, Long, Wang, and Jordan}{You
  et~al\mbox{.}}{2019}]%
        {you2019does}
\bibfield{author}{\bibinfo{person}{Kaichao You}, \bibinfo{person}{Mingsheng
  Long}, \bibinfo{person}{Jianmin Wang}, {and} \bibinfo{person}{Michael~I
  Jordan}.} \bibinfo{year}{2019}\natexlab{}.
\newblock \showarticletitle{How does learning rate decay help modern neural
  networks?}
\newblock \bibinfo{journal}{\emph{arXiv preprint arXiv:1908.01878}}
  (\bibinfo{year}{2019}).
\newblock


\bibitem[\protect\citeauthoryear{Yun, Han, Oh, Chun, Choe, and Yoo}{Yun
  et~al\mbox{.}}{2019}]%
        {yun2019cutmix}
\bibfield{author}{\bibinfo{person}{Sangdoo Yun}, \bibinfo{person}{Dongyoon
  Han}, \bibinfo{person}{Seong~Joon Oh}, \bibinfo{person}{Sanghyuk Chun},
  \bibinfo{person}{Junsuk Choe}, {and} \bibinfo{person}{Youngjoon Yoo}.}
  \bibinfo{year}{2019}\natexlab{}.
\newblock \showarticletitle{Cutmix: Regularization strategy to train strong
  classifiers with localizable features}. In
  \bibinfo{booktitle}{\emph{Proceedings of the IEEE/CVF International
  Conference on Computer Vision}}. \bibinfo{pages}{6023--6032}.
\newblock


\bibitem[\protect\citeauthoryear{Zagoruyko and Komodakis}{Zagoruyko and
  Komodakis}{2016}]%
        {zagoruyko2016wide}
\bibfield{author}{\bibinfo{person}{Sergey Zagoruyko} {and}
  \bibinfo{person}{Nikos Komodakis}.} \bibinfo{year}{2016}\natexlab{}.
\newblock \showarticletitle{Wide residual networks}.
\newblock \bibinfo{journal}{\emph{arXiv preprint arXiv:1605.07146}}
  (\bibinfo{year}{2016}).
\newblock


\bibitem[\protect\citeauthoryear{Zoph and Le}{Zoph and Le}{2016}]%
        {zoph2016neural}
\bibfield{author}{\bibinfo{person}{Barret Zoph} {and} \bibinfo{person}{Quoc~V
  Le}.} \bibinfo{year}{2016}\natexlab{}.
\newblock \showarticletitle{Neural architecture search with reinforcement
  learning}.
\newblock \bibinfo{journal}{\emph{arXiv preprint arXiv:1611.01578}}
  (\bibinfo{year}{2016}).
\newblock


\bibitem[\protect\citeauthoryear{Zoph, Vasudevan, Shlens, and Le}{Zoph
  et~al\mbox{.}}{2018}]%
        {zoph2018learning}
\bibfield{author}{\bibinfo{person}{Barret Zoph}, \bibinfo{person}{Vijay
  Vasudevan}, \bibinfo{person}{Jonathon Shlens}, {and} \bibinfo{person}{Quoc~V
  Le}.} \bibinfo{year}{2018}\natexlab{}.
\newblock \showarticletitle{Learning transferable architectures for scalable
  image recognition}. In \bibinfo{booktitle}{\emph{Proceedings of the IEEE
  Conference on Computer Vision and Pattern Recognition (CVPR)}}.
  \bibinfo{pages}{8697--8710}.
\newblock


\end{thebibliography}

\end{document}